\documentclass{article} 
\usepackage{iclr2026_conference,times}
\usepackage{graphicx}
\usepackage{booktabs}
\usepackage{algorithm}
\usepackage{algorithmic}
\usepackage{wrapfig}
\usepackage{listings}
\usepackage{xcolor}
\usepackage{multicol}
\usepackage{tcolorbox}
\usepackage{enumitem}

\usepackage{amsmath,amsfonts,bm}

\def\eqref#1{equation~\ref{#1}}

\def\1{\bm{1}}

\DeclareMathAlphabet{\mathsfit}{\encodingdefault}{\sfdefault}{m}{sl}
\SetMathAlphabet{\mathsfit}{bold}{\encodingdefault}{\sfdefault}{bx}{n}

\usepackage{hyperref}
\definecolor{myteal}{HTML}{38A6A5}
\definecolor{mynavy}{HTML}{1D6996}

\hypersetup{
    colorlinks=true,
    citecolor=mynavy,
    linkcolor=mynavy,
    urlcolor=mynavy,
    filecolor=black
}
\usepackage{url}

\tcbset{
  takeawaybox/.style={
    colback=myteal!3!white,
    colframe=myteal,
    fonttitle=\bfseries\small,
    title=Takeaways of the Experimental Section,
    boxrule=0.8pt,
    arc=3pt,
    left=4pt, right=4pt, top=3pt, bottom=3pt,
  }
}

\title{Three Steps at a Time: Learning \\ Representations from Action Sequences \\ in Contrastive RL}

\author{Michal Korniak$^*$ \\
ETH Zurich \\
\And
Kamil Dybek$^*$ \\
University of Warsaw \\
\And
Benjamin Eysenbach$^\dagger$ \\
Princeton University \\
\AND
Marco Bagatella$^\dagger$ \\
ETH Zurich \\ MPI IS Tubingen \\
\And
Michał Bortkiewicz$^\dagger$ \\
Warsaw University of Technology \\
IDEAS Research Institute
}

\iclrfinalcopy 
\begin{document}

\maketitle

\def\thefootnote{}\footnotetext{$^*$Equal contribution. $^\dagger$Equal advising. Correspondence to: \href{mailto:michael.korniak@gmail.com}{{michael.korniak@gmail.com}}.}
\def\thefootnote{\arabic{footnote}}

\begin{abstract}
While self-supervised approaches to reinforcement learning
have achieved strong results by learning representations of states and actions, a key open question is the time scale over which actions should be modeled. Departing from the standard formulation relying on single-step actions, we extend contrastive reinforcement learning (CRL), a prototypical self-supervised method, to operate over action chunks, and find that this
results in large, pervasive gains across established offline and online benchmarks: $+31.7\%$ and $+93.1\%$ across 18 and 11 environments respectively. While action-chunking-driven gains are generally explained through the ability to model non-Markovian, temporally extended policies, and to propagate unbiased multi-step returns, interestingly, we find that these arguments only partially apply to CRL.
Our empirical studies suggest that, in the context of CRL, an action chunk carries more information about the goal than a single action, measurably improving the critic's representations, and rendering the algorithm significantly more effective.
\end{abstract}

\section{Introduction}

Self-supervised reinforcement learning algorithms have gradually emerged as promising methods for learning diverse behavior with minimal supervision. 
Among them, contrastive reinforcement learning (CRL) \citep{eysenbach2022contrastive} tackles this challenge by learning
representations that encode the solution to a simple classification problem: \textit{will this goal be visited in the future, when starting from a given state and executing a single action?}

\begin{figure}[t]
    \centering
    \vspace{-12mm}
    \includegraphics[width=0.9\textwidth]{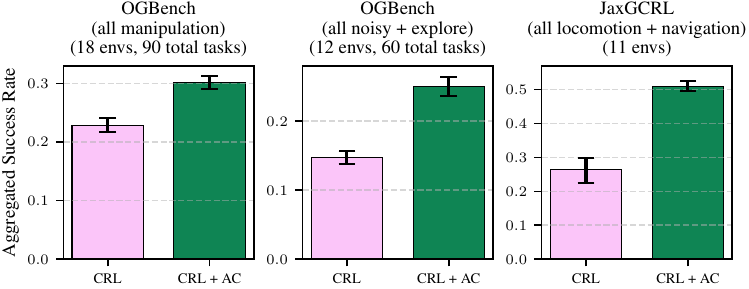}
    \caption{\textbf{Action chunking boosts goal-reaching performance.
    } We compare CRL + AC with a fixed action chunk length ($H {=} 3$) against standard CRL across 3 settings, reporting 95\% bootstrapped confidence intervals. CRL + AC consistently outperforms CRL, with average improvements of $+31.7\%$ on the OGBench manipulation suite, $+69.4\%$ on suboptimal OGBench datasets (\texttt{noisy} and \texttt{explore}), and $+93.1\%$ on online JaxGCRL locomotion and navigation tasks.
    }
    \label{fig:frontpage}
    \vspace{-2mm}
\end{figure}

In this work, we study whether stronger critic representations can be learned by conditioning this classification problem on \emph{sequences} of actions.
In line with prior work on action chunking \citep{zhaoLearningFineGrainedBimanual2023,liReinforcementLearningAction2026a}, we find that this minimal modification induces consistent and substantial performance gains across established benchmarks, spanning from offline manipulation to online locomotion and navigation across numerous tasks, as summarized in Figure~\ref{fig:frontpage}. 

However, we find that the standard explanations for \emph{why} action chunking helps only partially apply to CRL, and a synergy specific to CRL is at play in this setting.
On top of the ability to model non-Markovian data, and to express temporally consistent behavior \citep{liDecoupledQChunking2025}, we find that action chunking gives the critic more information about the goal than a single action, measurably improving its representations, and rendering the algorithm significantly more effective. Moreover, we find that conditioning on sequences of actions is particularly useful on noisy datasets, where individual actions are often weakly informative about future outcomes.

Finally, we observe that this phenomenon compounds with previously researched benefits of action chunking, and with the model scaling properties of CRL \citep{wang20261000}.
Our results suggest that explicit modeling of action sequences should be explored beyond the supervised or temporal-difference objectives for which it was originally proposed. Our contributions are: (i) we investigate a simple extension of CRL that conditions the critic on a chunk of actions rather than a single action; (ii) we empirically validate this approach across established offline and online benchmarks, and report increases in performance of $+31.7\%$ and $+93.1\%$ respectively; (iii) we study the causes of these improvements and find that, beyond the known benefits of action chunking, it synergizes specifically with CRL: conditioning the critic on an action chunk gives it more information about the goal than a single action, producing measurably better critic representations.

\section{Related Work}
\paragraph{Goal-conditioned RL and action representations.}
Our study focuses on the goal-conditioned RL (GCRL) setup in which an agent learns to achieve various tasks through conditioning on goals~\citep{kaelblingLearningAchieveGoals1993,schaul2015universal}. In this setup, both the policy and the critic are conditioned on an additional goal input, such as proprioception state configuration~\citep{ghoshLearningActionableRepresentations2018}, a visual scene representation~\citep{nairVisualReinforcementLearning2018}, or a language instruction~\citep{myersTemporalRepresentationAlignment2025,hermannGroundedLanguageLearning2017,bahdanauLearningUnderstandGoal2019}.
In reinforcement learning (RL), and particularly in GCRL, representations of states and goals can be organized by their consequences for control rather than perceptual similarity. For example, actionable representations place two goals close together when they induce similar action distributions~\citep{ghoshLearningActionableRepresentations2018}; bisimulation-based representations group states that produce equivalent rewards and transition distributions under each action~\citep{GIVAN2003163,ferns2012metrics,DBLP:conf/icml/GeladaKBNB19,DBLP:conf/iclr/0001MCGL21,DBLP:conf/icml/Hansen-Estruch022}; and temporal-distance representations organize state--goal pairs by their reachability or the number of actions required to connect them~\citep{dayan1993improving, wangOptimalGoalReachingReinforcement2023,myersLearningTemporalDistances2024,steccanella2022state}.

\paragraph{Contrastive Learning and Contrastive RL.}
Contrastive Learning is a widely used method to learn rich representations from unlabeled data across vision~\citep{chenSimpleFrameworkContrastive2020,wangUnderstandingContrastiveRepresentation2022,yehDecoupledContrastiveLearning2021,oordRepresentationLearningContrastive2019} and other modalities~\citep{radfordLearningTransferableVisual2021,zhaiSigmoidLossLanguage2023}. In our work, we focus on temporal contrastive learning~\citep{oordRepresentationLearningContrastive2019}, specifically Contrastive Reinforcement Learning \nopagebreak(CRL)~\citep{eysenbach2022contrastive}, which casts GCRL as a representation-learning problem in which the objective of the value function is to distinguish states likely to occur in the near future from those unlikely to occur.
Recent work has shown promising results in terms of model scalability~\citep{zheng2024stabilizing,wang20261000}, and sought to explain how learned representations drive exploration~\citep{liuSingleGoalAll2024,bastankhahDemystifyingMechanismsEmergent2025}; we find that controlling action representations also provides significant gains. In parallel, another line of work has addressed the asymmetry of temporal distances by introducing quasimetric formulations~\citep{myersLearningTemporalDistances2024,myersTemporalRepresentationAlignment2025}. Recent methods reduce reliance on static trajectory-level context when relating state-action pairs to goals~\citep{ziarkoContrastiveRepresentationsTemporal2025}. We build on this view, arguing that what the critic learns about the goal depends not only on the agent's state but also on the informativeness of its actions.

\paragraph{Action Chunking.}
Action chunking (AC)~\citep{zhaoLearningFineGrainedBimanual2023} parameterizes the policy’s output as a sequence of $H$ atomic actions from the original action space, $(a_t,\ldots,a_{t+H-1}) \in \mathcal{A}^H$, rather than as a single action. At the environment level, each control input remains an atomic action in $\mathcal{A}$. Under open-loop execution, however, the policy commits to a predicted chunk $a_{t:t+H-1}\in\mathcal{A}^H$ as a single temporally extended decision, whereas closed-loop variants execute only a couple of first actions before predicting a new chunk from the subsequent observations~\citep{zhaoLearningFineGrainedBimanual2023,DBLP:conf/iclr/LiuHX0DF25,liReinforcementLearningAction2026a}.
\looseness=-1 While the idea of temporally-extended action is not new~\citep{randlovLearningMacroActionsReinforcement1998,suttonMDPsSemiMDPsFramework1999}, the action chunking formulation has recently gained traction  for mitigating control latency in real-world deployments~\citep{blackRealTimeExecutionAction2025,blackTrainingTimeActionConditioning2025} and as an effective way to learn from demonstration datasets with multimodal action distributions~\citep{zhaoLearningFineGrainedBimanual2023,chiVisuomotorPolicyLearning}. Beyond these practical advantages, prior work has studied how open-loop execution of action chunks can reduce the effective decision horizon~\citep{zhaoLearningFineGrainedBimanual2023} and improve behavioral consistency~\citep{zhangActionChunkingExploratory2025}, but also might result in reduced reactivity~\citep{liuBidirectionalDecodingImproving2025}. In this work, we study action chunking in the context of GCRL, where it has been shown to effectively propagate action values, handle non-Markovian data~\citep{liReinforcementLearningAction2026a}, and promote temporally coherent behavior~\citep{shinAdaptiveActionChunking2026,liDecoupledQChunking2025}. As discussed later, we find that action chunking displays a special synergy with CRL, as sequences of actions carry more information about the intended goal than isolated atomic actions.
\section{Background}
We model the environment as a reward-free Markov Decision Process $\mathcal{M} = (\mathcal{S}, \mathcal{A}, P, \mu_0, \gamma)$, where $\mathcal{S}$ and $\mathcal{A}$ are state and action spaces, $P: \mathcal{S} \times \mathcal{A} \to \Delta(\mathcal{S})$ is a stochastic transition kernel, $\mu_0 \in \Delta(\mathcal{S})$ is an initial state distribution and $\gamma \in [0, 1)$ is a discount factor.
The goal space coincides with the state space: $\mathcal{G}:=\mathcal{S}$.
A stationary policy is defined as a mapping from state-goal pairs to action distributions $\pi: \mathcal{S} \times \mathcal{G} \to \Delta(\mathcal{A})$.
For a given policy $\pi$ conditioned on goal $g$ and starting from a state-action pair $(s,a)$, we can model the probability of visiting future state $s_f$ as
\begin{equation}
    p^{\pi(\cdot \mid \cdot, g)}(s_f \mid s,a) := (1-\gamma)\sum_{t\geq0}\gamma^t\mathbb{P}^{\pi(\cdot \mid \cdot, g)}(s_t = s_f \mid s_0=s, a_0=a),
\end{equation}

where the MDP unrolls through $a_t \sim \pi(s_t, g)$ and $s_t \sim P(s_{t-1}, a_{t-1})$, and $\mathbb{P}^{\pi(\cdot \mid \cdot, g)}(s_t = s_f \mid s_0=s, a_0=a)$ denotes the $t$-step transition probabilities.
This quantity is the \textit{successor measure}~\citep{dayan1993improving,blier2021learning}. The successor measure is equivalent to the Q-function for a goal-conditioned reward $R(s,g)=\mathbf{1}_{s=g}$:
\begin{equation*}
Q^\pi_g(s,a) = \mathbb{E}_{\pi, P} \sum_{t\geq0}\gamma^t R(s_t,g) = \frac{p^\pi(g|s,a)}{1-\gamma}.
\end{equation*}
For some goal distribution $\mu_g \in \Delta(\mathcal{G})$ we aim to find a policy that maximize the likelihood of reaching the goal (i.e., maximizes its Q-function):
%
\begin{equation}
    \arg \max_{\pi(\cdot \mid \cdot, \cdot)} \mathbb{E}_{\substack{g \sim \mu_g \\ s_0 \sim \mu_0 \\ a_0 \sim \pi}} p^{\pi(\cdot \mid \cdot, g)}(s_f = g \mid s_0, a_0).
\end{equation}
%

CRL recasts goal-conditioned RL as a contrastive representation learning problem, training encoders $\phi: \mathcal{S \times A} \to \mathbb{R}^d$
and $\psi: \mathcal{G} \to \mathbb{R}^d$ such that their inner product $f(s,a,g)=\phi(s,a)^\top\psi(g)$ recovers a log-probability ratio between a positive and a negative distribution.
In practice, given a state $s_t$, positive samples are future states from the same trajectory $s_{t+\Delta}$ where $\Delta$ is drawn geometrically (thus following the successor measure $p^\beta$ embedded in the dataset), and negative samples are uniformly sampled from the dataset (thus recovering the marginal of $p^\beta$). We can then train the representation to discriminate samples from the positive and negative distribution:
\begin{equation}
    \mathcal{L}(\phi, \psi) = \mathbb{E}_{\substack{(s,a) \sim \mu_0 \\ g^+ \sim p^\beta(g \mid s, a) \\ g^- \sim p^\beta(g)}} \Big[ \log \sigma\!\left(\phi(s,a)^\top \psi(g^+)\right) + \log\!\left(1 - \sigma\!\left(\phi(s,a)^\top \psi(g^-)\right)\right) \Big].
\end{equation}

This objective can be shown to lower bound the mutual information $I(s,a;g)$ \citep{eysenbach2022contrastive}.
At its optimum, the exponentiated dot products between learned representations are equal up to a constant to Q-values:
\begin{equation}
    \exp(\phi^\star(s,a)^\top \psi^\star(g)) = \frac{p^\beta(g|s,a)}{p^\beta(g)} = \frac{Q^\beta_g(s,a)}{Z_g(s)},
\end{equation}

where $Z_g(s)$ is a constant that is action independent and thus does not need to be estimated.
Thus the representations can be used to rank actions at each state.
A policy can be trained to maximize these dot-products with the following loss:
\begin{equation}
    \mathcal{L}(\pi) = -\,\mathbb{E}_{\substack{s, g \sim p^\beta(g) \\ a \sim \pi(\cdot |s,g)}} \big[\, \phi(s, a)^\top \psi(g) \,\big].
\end{equation}

\section{Action-Chunked Contrastive RL}
We study a simple extension of CRL that conditions the critic on a sequence of $H$ future actions — an action chunk — rather than a single action. We refer to it as CRL + AC.

\paragraph{Training Time: Action-Chunked Contrastive Objective.} We condition the critic on a sequence of $H$ consecutive actions, $a_{1:H} = (a_1, \ldots, a_H) \in \mathcal{A}^H$, rather than a single action:
\begin{equation}
    f(s, a_{1:H}, g) = \phi(s, a_{1:H})^\top \psi(g).
\end{equation}
The positive goal $g$ is sampled geometrically from future states along the trajectory, exactly as in standard CRL and independently of $H$ (i.e., $g$ need not be the state $H$ steps ahead). The actor $\pi$ is similarly modified to output a full action chunk, i.e., $\pi(s, g) = a_{1:H} \in \mathcal{A}^H$, and is trained via the same policy improvement objective as CRL:
\begin{equation}
    \mathcal{L}(\pi) = -\,\mathbb{E}_{\substack{s, g \sim p^\beta(g) \\ a_{1:H} \sim \pi(\cdot |s,g)}} \big[\, f(s, a_{1:H}, g) \,\big].
\end{equation}
Aside from the critic's input and the actor's output, the training pipeline is largely unchanged. The replay buffer stores trajectories exactly as before; the only difference is at sampling time, where $a_{1:H}$ is assembled by concatenating $H$ consecutive stored actions and the goal  $g$ is drawn geometrically from future states along the trajectory.

\paragraph{Test-time: Chunk Execution and Goal Conditioning.} During inference, the policy predicts an action chunk $a_{1:H}$ but need not execute all of it before requerying. At a replanning interval $H_{\text{exec}} \le H$, the policy executes only the first $H_{\text{exec}}$ actions of each predicted chunk before requerying the policy: $H_{\text{exec}}{=}H$ recovers full open-loop execution (the chunk is run to completion with no intermediate replanning), while $H_{\text{exec}}{=}1$ replans at every step. Full open-loop execution ($H_{\text{exec}}{=}H$) is our default for the main results; we
ablate the replanning interval in Section~\ref{sec:experiments}, and replan at every step
($H_{\text{exec}}{=}1$) in our analysis (Section~\ref{subsec:why_does_ac_help}) to isolate the effect of open-loop execution.

\paragraph{Implementation.} The modification to standard CRL is minimal. The replay buffer stores trajectories in the standard way; action chunks and goals are constructed at sampling time — $a_{1:H}$ by concatenating $H$ consecutive actions, and $g$ by sampling future states along the trajectory, geometrically sampled from $s$. 
Architecturally, the critic and actor networks are the only components that change, and both remain multilayer perceptrons: the critic takes the flattened chunk $a_{1:H}$ as input instead of a single action, and the actor outputs $a_{1:H}$ instead of a single action. Algorithm~\ref{alg:ac-crl} shows the resulting critic and actor losses, adapted from \citet{eysenbach2022contrastive}.

Since only the critic's input and the actor's output change, CRL + AC is agnostic to how the policy is extracted from the critic: any extraction procedure (e.g., DDPG-style objective~\citep{barth2018distributed}, AWR~\citep{peng2019advantageweightedregressionsimplescalable}, or FQL~\citep{park2025flowqlearning}) can be applied on top of the chunked critic. In our experiments (Section~\ref{sec:experiments}) we use FQL in the offline setting and a DDPG-style actor with entropy regularization in the online setting~\citep{williams1991function, heess2015learning}.

\section{Experiments}
\label{sec:experiments}
In this section we will first evaluate action chunking in CRL thoroughly across two settings: online and offline (Section \ref{sec:exp_results}). We then ask why chunking helps CRL specifically, identifying a representational effect beyond the standard explanations (Section~\ref{subsec:why_does_ac_help}). Section~\ref{sec:exp_ablations} ablates design choices.

\begin{figure}[t]
    \centering
    \vspace{-12mm}
    \includegraphics[width=1.0\textwidth]{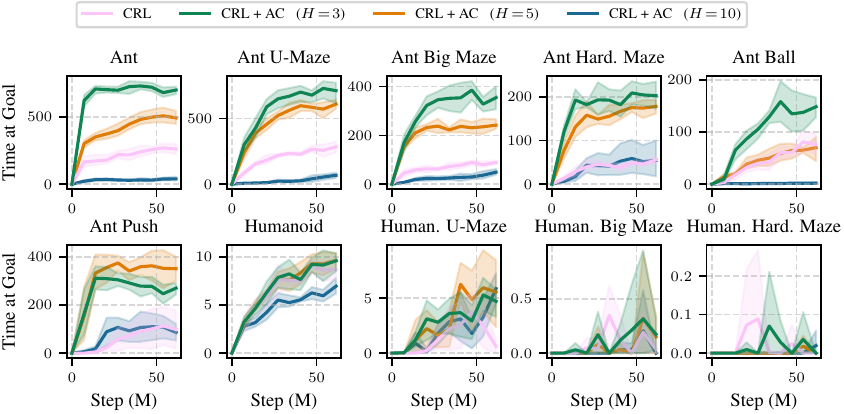}
    \caption{\textbf{Online locomotion and navigation results.} We report time-at-goal learning curves of CRL and CRL + AC with action chunk length $H \in \{3, 5, 10 \}$ on 10 JaxGCRL locomotion and navigation tasks, with 95\% bootstrapped confidence intervals. Time-at-goal captures both task success and behavioral quality, rewarding policies that reach and stably maintain the goal~\citep{bortkiewicz2025accelerating}. Overall, CRL + AC with $H \in \{3, 5\}$ consistently and substantially outperforms CRL, with $H {=}3$ being the best or near-best chunk length across the majority of tasks, improving the performance by $+111.2\%$ on average when all 11 JaxGCRL environments are considered.}
    \label{fig:learning_curves_10envs}
\end{figure}

\paragraph{Experimental setup.}
In the online setting, the agent collects its own experience while learning, so any effect of action chunking is mediated jointly through representation learning and exploration. In the offline setting, we evaluate CRL + AC on datasets of varying data quality, from near-expert demonstrations to noisy and exploratory data, isolating the effect of data quality on performance.

In both regimes, we keep the original algorithm and hyperparameters unchanged unless otherwise specified, modifying only the contrastive critic's input and the actor's output to operate on chunked actions. Training protocols are kept unchanged across all experiments. We report 95\% stratified confidence intervals computed with RLiable~\citep{agarwal2021deep}. Full details on hyperparameters, evaluation protocols, detailed figure descriptions, additional ablations, and learning curves are provided in the Appendix~\ref{appendix}-\ref{app:learning_curves}. Code is available at \url{https://github.com/M-Korniak/action-chunked-contrastive-rl}.

\subsection{Results}
\label{sec:exp_results}

\paragraph{Online Reinforcement Learning.}
We first investigate whether action chunking improves CRL in the online setting, and whether the improvement holds broadly across a diverse range of tasks. We evaluate on JaxGCRL~\citep{bortkiewicz2025accelerating}, a goal-conditioned RL framework, using its broad range of locomotion and navigation tasks with the default network architecture and training protocol. We run experiments for standard CRL and CRL + AC with action chunk lengths $H \in \{3, 5, 10, 15\}$. The actor's target entropy is scaled to the dimensionality of the action chunk, rather than that of an individual action, ensuring consistent entropy regularization under the expanded action space. We report two metrics: success rate and time-at-goal, the latter adopted from JaxGCRL~\citep{bortkiewicz2025accelerating} and used in prior work~\citep{wang20261000}. 

\looseness=-1 Figure~\ref{fig:learning_curves_10envs} demonstrates that action chunking improves performance on online locomotion and navigation tasks. Averaged across all evaluated online tasks, it increases time-at-goal by $+111.2\%$ and success rate by $+93.1\%$ (Appendix~\ref{app:jaxgcrl_learning_curves}). Action chunking thus improves CRL in the online setting, and the gains are consistent across the task suite rather than confined to individual environments.

\begin{table}[t]
\centering
\vspace{-12mm}
\scalebox{0.85}{
\begin{tabular}{lrr@{\quad}rrrrr}
\toprule
OGBench Environment & CRL & CRL + AC & {\color{gray!80}GCBC} & {\color{gray!80}GCIVL} & {\color{gray!80}GCIQL} & {\color{gray!80}QRL} & {\color{gray!80}HIQL} \\
\midrule
\texttt{cube-single-play-v0} & $74$ {\tiny $\pm 4$} & $\mathbf{81}$ {\tiny $\pm \mathbf{3}$} & {\color{gray!80}$6$ {\tiny $\pm 2$}} & {\color{gray!80}$53$ {\tiny $\pm 4$}} & {\color{gray!80}$68$ {\tiny $\pm 6$}} & {\color{gray!80}$5$ {\tiny $\pm 1$}} & {\color{gray!80}$15$ {\tiny $\pm 3$}} \\
\texttt{cube-double-play-v0} & $\mathbf{29}$ {\tiny $\pm \mathbf{4}$} & $25$ {\tiny $\pm 5$} & {\color{gray!80}$1$ {\tiny $\pm 1$}} & {\color{gray!80}$36$ {\tiny $\pm 3$}} & {\color{gray!80}$40$ {\tiny $\pm 5$}} & {\color{gray!80}$1$ {\tiny $\pm 0$}} & {\color{gray!80}$6$ {\tiny $\pm 2$}} \\
\texttt{cube-triple-play-v0} & $\mathbf{12}$ {\tiny $\pm \mathbf{8}$} & $\mathbf{5}$ {\tiny $\pm \mathbf{2}$} & {\color{gray!80}$1$ {\tiny $\pm 1$}} & {\color{gray!80}$1$ {\tiny $\pm 0$}} & {\color{gray!80}$3$ {\tiny $\pm 1$}} & {\color{gray!80}$0$ {\tiny $\pm 0$}} & {\color{gray!80}$3$ {\tiny $\pm 1$}} \\
\texttt{cube-quadruple-play-v0} & $\mathbf{0}$ {\tiny $\pm \mathbf{0}$} & $\mathbf{0}$ {\tiny $\pm \mathbf{0}$} & {\color{gray!80}$0$ {\tiny $\pm 0$}} & {\color{gray!80}$0$ {\tiny $\pm 0$}} & {\color{gray!80}$0$ {\tiny $\pm 0$}} & {\color{gray!80}$0$ {\tiny $\pm 0$}} & {\color{gray!80}$0$ {\tiny $\pm 0$}} \\
\cmidrule{1-8}
\textit{Cube (play avg.)} & $\mathbf{29}$ {\tiny $\pm \mathbf{2}$} & $\mathbf{28}$ {\tiny $\pm \mathbf{1}$} & {\color{gray!80}$2$ {\tiny $\pm 0$}} & {\color{gray!80}$22$ {\tiny $\pm 1$}} & {\color{gray!80}$28$ {\tiny $\pm 1$}} & {\color{gray!80}$1$ {\tiny $\pm 0$}} & {\color{gray!80}$6$ {\tiny $\pm 1$}} \\
\midrule
\texttt{cube-single-noisy-v0} & $\mathbf{86}$ {\tiny $\pm \mathbf{4}$} & $74$ {\tiny $\pm 3$} & {\color{gray!80}$8$ {\tiny $\pm 3$}} & {\color{gray!80}$71$ {\tiny $\pm 9$}} & {\color{gray!80}$99$ {\tiny $\pm 1$}} & {\color{gray!80}$25$ {\tiny $\pm 6$}} & {\color{gray!80}$41$ {\tiny $\pm 6$}} \\
\texttt{cube-double-noisy-v0} & $10$ {\tiny $\pm 1$} & $\mathbf{25}$ {\tiny $\pm \mathbf{3}$} & {\color{gray!80}$1$ {\tiny $\pm 1$}} & {\color{gray!80}$14$ {\tiny $\pm 3$}} & {\color{gray!80}$23$ {\tiny $\pm 3$}} & {\color{gray!80}$3$ {\tiny $\pm 1$}} & {\color{gray!80}$2$ {\tiny $\pm 1$}} \\
\texttt{cube-triple-noisy-v0} & $\mathbf{2}$ {\tiny $\pm \mathbf{1}$} & $\mathbf{3}$ {\tiny $\pm \mathbf{2}$} & {\color{gray!80}$1$ {\tiny $\pm 1$}} & {\color{gray!80}$9$ {\tiny $\pm 1$}} & {\color{gray!80}$2$ {\tiny $\pm 1$}} & {\color{gray!80}$1$ {\tiny $\pm 0$}} & {\color{gray!80}$2$ {\tiny $\pm 1$}} \\
\texttt{cube-quadruple-noisy-v0} & $\mathbf{0}$ {\tiny $\pm \mathbf{0}$} & $\mathbf{0}$ {\tiny $\pm \mathbf{0}$} & {\color{gray!80}$0$ {\tiny $\pm 0$}} & {\color{gray!80}$0$ {\tiny $\pm 0$}} & {\color{gray!80}$0$ {\tiny $\pm 0$}} & {\color{gray!80}$0$ {\tiny $\pm 0$}} & {\color{gray!80}$0$ {\tiny $\pm 0$}} \\
\cmidrule{1-8}
\textit{Cube (noisy avg.)} & $\mathbf{24}$ {\tiny $\pm \mathbf{1}$} & $\mathbf{26}$ {\tiny $\pm \mathbf{1}$} & {\color{gray!80}$2$ {\tiny $\pm 1$}} & {\color{gray!80}$23$ {\tiny $\pm 2$}} & {\color{gray!80}$31$ {\tiny $\pm 0$}} & {\color{gray!80}$7$ {\tiny $\pm 1$}} & {\color{gray!80}$11$ {\tiny $\pm 1$}} \\
\midrule
\texttt{puzzle-3x3-play-v0} & $\mathbf{28}$ {\tiny $\pm \mathbf{3}$} & $21$ {\tiny $\pm 7$} & {\color{gray!80}$2$ {\tiny $\pm 0$}} & {\color{gray!80}$6$ {\tiny $\pm 1$}} & {\color{gray!80}$95$ {\tiny $\pm 1$}} & {\color{gray!80}$1$ {\tiny $\pm 0$}} & {\color{gray!80}$12$ {\tiny $\pm 2$}} \\
\texttt{puzzle-4x4-play-v0} & $58$ {\tiny $\pm 5$} & $\mathbf{62}$ {\tiny $\pm \mathbf{3}$} & {\color{gray!80}$0$ {\tiny $\pm 0$}} & {\color{gray!80}$13$ {\tiny $\pm 2$}} & {\color{gray!80}$26$ {\tiny $\pm 3$}} & {\color{gray!80}$0$ {\tiny $\pm 0$}} & {\color{gray!80}$7$ {\tiny $\pm 2$}} \\
\texttt{puzzle-4x5-play-v0} & $18$ {\tiny $\pm 2$} & $\mathbf{19}$ {\tiny $\pm \mathbf{0}$} & {\color{gray!80}$0$ {\tiny $\pm 0$}} & {\color{gray!80}$7$ {\tiny $\pm 1$}} & {\color{gray!80}$14$ {\tiny $\pm 1$}} & {\color{gray!80}$0$ {\tiny $\pm 0$}} & {\color{gray!80}$4$ {\tiny $\pm 1$}} \\
\texttt{puzzle-4x6-play-v0} & $\mathbf{16}$ {\tiny $\pm \mathbf{1}$} & $15$ {\tiny $\pm 1$} & {\color{gray!80}$0$ {\tiny $\pm 0$}} & {\color{gray!80}$10$ {\tiny $\pm 2$}} & {\color{gray!80}$12$ {\tiny $\pm 1$}} & {\color{gray!80}$0$ {\tiny $\pm 0$}} & {\color{gray!80}$3$ {\tiny $\pm 1$}} \\
\cmidrule{1-8}
\textit{Puzzle (play avg.)} & $\mathbf{29}$ {\tiny $\pm \mathbf{4}$} & $\mathbf{28}$ {\tiny $\pm \mathbf{4}$} & {\color{gray!80}$0$ {\tiny $\pm 0$}} & {\color{gray!80}$9$ {\tiny $\pm 0$}} & {\color{gray!80}$37$ {\tiny $\pm 0$}} & {\color{gray!80}$0$ {\tiny $\pm 0$}} & {\color{gray!80}$7$ {\tiny $\pm 0$}} \\
\midrule
\texttt{puzzle-3x3-noisy-v0} & $35$ {\tiny $\pm 3$} & $\mathbf{56}$ {\tiny $\pm \mathbf{3}$} & {\color{gray!80}$1$ {\tiny $\pm 0$}} & {\color{gray!80}$42$ {\tiny $\pm 19$}} & {\color{gray!80}$94$ {\tiny $\pm 3$}} & {\color{gray!80}$0$ {\tiny $\pm 0$}} & {\color{gray!80}$51$ {\tiny $\pm 11$}} \\
\texttt{puzzle-4x4-noisy-v0} & $1$ {\tiny $\pm 1$} & $\mathbf{42}$ {\tiny $\pm \mathbf{3}$} & {\color{gray!80}$0$ {\tiny $\pm 0$}} & {\color{gray!80}$20$ {\tiny $\pm 3$}} & {\color{gray!80}$29$ {\tiny $\pm 7$}} & {\color{gray!80}$0$ {\tiny $\pm 0$}} & {\color{gray!80}$16$ {\tiny $\pm 4$}} \\
\texttt{puzzle-4x5-noisy-v0} & $4$ {\tiny $\pm 1$} & $\mathbf{16}$ {\tiny $\pm \mathbf{1}$} & {\color{gray!80}$0$ {\tiny $\pm 0$}} & {\color{gray!80}$19$ {\tiny $\pm 0$}} & {\color{gray!80}$19$ {\tiny $\pm 0$}} & {\color{gray!80}$0$ {\tiny $\pm 0$}} & {\color{gray!80}$5$ {\tiny $\pm 1$}} \\
\texttt{puzzle-4x6-noisy-v0} & $4$ {\tiny $\pm 2$} & $\mathbf{15}$ {\tiny $\pm \mathbf{2}$} & {\color{gray!80}$0$ {\tiny $\pm 0$}} & {\color{gray!80}$17$ {\tiny $\pm 2$}} & {\color{gray!80}$18$ {\tiny $\pm 2$}} & {\color{gray!80}$0$ {\tiny $\pm 0$}} & {\color{gray!80}$2$ {\tiny $\pm 1$}} \\
\cmidrule{1-8}
\textit{Puzzle (noisy avg.)} & $9$ {\tiny $\pm 3$} & $\mathbf{34}$ {\tiny $\pm \mathbf{2}$} & {\color{gray!80}$0$ {\tiny $\pm 0$}} & {\color{gray!80}$23$ {\tiny $\pm 3$}} & {\color{gray!80}$40$ {\tiny $\pm 1$}} & {\color{gray!80}$0$ {\tiny $\pm 0$}} & {\color{gray!80}$18$ {\tiny $\pm 2$}} \\
\midrule
\texttt{scene-play-v0} & $33$ {\tiny $\pm 9$} & $\mathbf{54}$ {\tiny $\pm \mathbf{2}$} & {\color{gray!80}$5$ {\tiny $\pm 1$}} & {\color{gray!80}$42$ {\tiny $\pm 4$}} & {\color{gray!80}$51$ {\tiny $\pm 4$}} & {\color{gray!80}$5$ {\tiny $\pm 1$}} & {\color{gray!80}$38$ {\tiny $\pm 3$}} \\
\texttt{scene-noisy-v0} & $17$ {\tiny $\pm 4$} & $\mathbf{30}$ {\tiny $\pm \mathbf{3}$} & {\color{gray!80}$1$ {\tiny $\pm 1$}} & {\color{gray!80}$26$ {\tiny $\pm 5$}} & {\color{gray!80}$26$ {\tiny $\pm 2$}} & {\color{gray!80}$9$ {\tiny $\pm 2$}} & {\color{gray!80}$25$ {\tiny $\pm 4$}} \\
\cmidrule{1-8}
\textit{Scene (avg.)} & $24$ {\tiny $\pm 7$} & $\mathbf{44}$ {\tiny $\pm \mathbf{6}$} & {\color{gray!80}$3$ {\tiny $\pm 0$}} & {\color{gray!80}$34$ {\tiny $\pm 2$}} & {\color{gray!80}$38$ {\tiny $\pm 1$}} & {\color{gray!80}$7$ {\tiny $\pm 1$}} & {\color{gray!80}$31$ {\tiny $\pm 1$}} \\
\midrule
\texttt{antmaze-medium-explore-v0} & $\mathbf{5}$ {\tiny $\pm \mathbf{3}$} & $\mathbf{4}$ {\tiny $\pm \mathbf{4}$} & {\color{gray!80}$2$ {\tiny $\pm 1$}} & {\color{gray!80}$19$ {\tiny $\pm 3$}} & {\color{gray!80}$13$ {\tiny $\pm 2$}} & {\color{gray!80}$1$ {\tiny $\pm 1$}} & {\color{gray!80}$37$ {\tiny $\pm 10$}} \\
\texttt{antmaze-large-explore-v0} & $\mathbf{0}$ {\tiny $\pm \mathbf{0}$} & $\mathbf{0}$ {\tiny $\pm \mathbf{0}$} & {\color{gray!80}$0$ {\tiny $\pm 0$}} & {\color{gray!80}$10$ {\tiny $\pm 3$}} & {\color{gray!80}$0$ {\tiny $\pm 0$}} & {\color{gray!80}$0$ {\tiny $\pm 0$}} & {\color{gray!80}$4$ {\tiny $\pm 5$}} \\
\texttt{antmaze-teleport-explore-v0} & $19$ {\tiny $\pm 3$} & $\mathbf{25}$ {\tiny $\pm \mathbf{3}$} & {\color{gray!80}$2$ {\tiny $\pm 1$}} & {\color{gray!80}$32$ {\tiny $\pm 2$}} & {\color{gray!80}$7$ {\tiny $\pm 3$}} & {\color{gray!80}$2$ {\tiny $\pm 2$}} & {\color{gray!80}$34$ {\tiny $\pm 15$}} \\
\cmidrule{1-8}
\textit{AntMaze (avg.)} & $\mathbf{8}$ {\tiny $\pm \mathbf{2}$} & $\mathbf{9}$ {\tiny $\pm \mathbf{4}$} & {\color{gray!80}$1$ {\tiny $\pm 0$}} & {\color{gray!80}$20$ {\tiny $\pm 1$}} & {\color{gray!80}$7$ {\tiny $\pm 1$}} & {\color{gray!80}$1$ {\tiny $\pm 0$}} & {\color{gray!80}$25$ {\tiny $\pm 3$}} \\
\bottomrule
\end{tabular}
}
\caption{\looseness = -1 \textbf{Offline manipulation and suboptimal data results.} We summarize per-environment success rate on OGBench~\citep{park2025ogbench} with action chunk length $H {=} 3$. CRL + AC improves over CRL by $+31.7\%$ on manipulation tasks and $+69.4\%$ on noisy and exploratory dataset variants, without degrading performance on any task group. Gains are most pronounced on noisy datasets, particularly \textit{Puzzle} and \textit{Scene}. \textit{Italic rows}: group aggregate computed via RLiable with 95\% bootstrapped CIs over seeds ${\times}$ tasks. Baselines are copied from the OGBench~\citep{park2025ogbench} paper for reference, with CIs obtained via parametric bootstrap from reported means and standard deviations. Bold denotes performance within the 95\% CI of the best among CRL and CRL + AC.}
\label{tab:ogbench_results_means}
\vspace{-4mm}
\end{table}

\paragraph{Offline Reinforcement Learning.} Having established that action chunking helps online, we next ask whether the benefit transfers to the offline setting, where exploration plays no role and data quality is fixed in advance, and whether it depends on that data quality. We evaluate CRL and CRL + AC (with action chunk length $H {=} 3$) on OGBench~\citep{park2025ogbench}, a standard offline goal-conditioned RL benchmark. We train on three dataset types: \texttt{play}, \texttt{noisy}, and \texttt{explore}, primarily across manipulation tasks. To better align with current state-of-the-art offline RL methods, we combine action chunking with a flow matching policy~\citep{lipman2022flow} and FQL~\citep{park2025flowqlearning} policy extraction. Since FQL is sensitive to the $\alpha$ parameter, which balances Q-improvement and behavioral cloning losses, we sweep $\alpha \in \{1, 3, 10\}$ for each environment and for both CRL and CRL + AC, selecting the best-performing value. We additionally compare a smaller-than-usual discount factor of $\gamma = 0.95$ against the standard $\gamma = 0.99$ for both CRL and CRL + AC, and find $\gamma = 0.95$ improves both; we therefore adopt it throughout (Appendix~\ref{app:gamma_ablation}).

Table~\ref{tab:ogbench_results_means} presents the results. Action chunking yields consistent improvements across all settings: an average of +31.7\% on manipulation tasks overall, with gains increasing to +69.4\% on the suboptimal \texttt{noisy} and \texttt{explore} dataset variants. These results confirm that the benefit of action chunking is not limited to the online setting. Performance on \texttt{play} environments remains on par with standard CRL, showing that the gains on noisy data do not come at the cost of performance on clean data.

\begin{figure}[t]
    \centering  
    \vspace{-12mm}
    \includegraphics[width=0.9\textwidth]{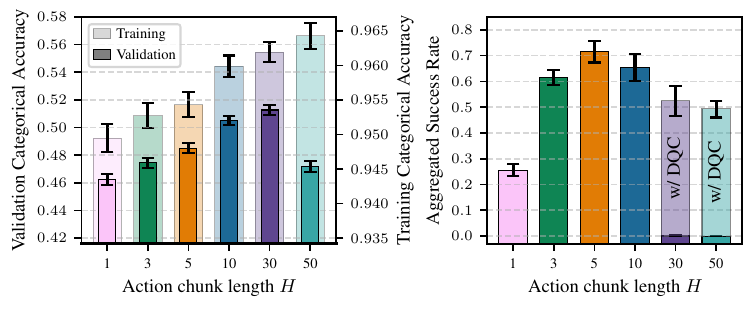}
    \vspace{-4mm}
    \caption{\textbf{Action chunking adds information about the goal; performance peaks at moderate chunk length.} Results on \texttt{cube-single-noisy-v0} ($\sigma {=} 1$). \textit{(Left)} Validation categorical accuracy (dark bars, left axis) increases with action chunk
length $H$ as the critic recovers more information about the goal. The thick line on
the x-axis marks the state-only critic's accuracy ($H{=}0$, no action conditioning;
Eq.~\ref{eq:value_critic}), so each bar's height shows the accuracy gained over the
state-only baseline. Accuracy peaks at $H{=}30$; at $H{=}50$ it drops even as
\emph{training} accuracy (light bars, left axis) continues to rise, indicating that
the long-chunk critic overfits. \textit{(Right)} Aggregated success rate peaks at a short chunk
    length ($H{=}5$) and collapses to near zero by $H{=}30$ and $H{=}50$. Distilling each long-chunk critic into a short-chunk
    ($H_\pi{=}5$) critic, and extracting policy from it (light bars, w/ DQC) recovers success to ${\sim}50\%$.
    All methods execute a single action and replan at each step ($H_{\text{exec}}{=}1$), so the collapse does not stem from open-loop execution, but from the difficulty of extracting an action from a long-chunk critic. We provide the full
per-$H$ learning curves and the DQC training curves in
Figure~\ref{fig:all_dqc_learning_curves} (Appendix~\ref{app:dqc}). Error bars show 95\% bootstrapped confidence intervals.}
\label{fig:why_ac_helps}
\vspace{-2mm}
\end{figure}

\subsection{Why does action chunking help?}
\label{subsec:why_does_ac_help}

\vspace{-4pt}
\looseness=-1 Having evaluated CRL with action chunking, we now turn to interpreting the sources of the performance gains. Prior work has put forth three explanations for why action chunking has proven useful for prior RL algorithms~\citep{liReinforcementLearningAction2026a}:
\vspace{-2mm}
\begin{enumerate}[noitemsep]
    \item the ability to model non-Markovian demonstrations;
    \item unbiased $H$-step returns for value estimation; and
    \item temporally extended exploration/exploitation.
\end{enumerate}
\vspace{-2mm}

 As we will show in the experiments below, the effectiveness of action chunking in CRL may not be explained by these arguments alone, and a mechanism particular to CRL is at play. Unlike most of the methods that have been reported to benefit from action chunking, CRL is strongly based on mutual information between state-action pairs and future outcomes. We thus put forth a new explanation for why action chunking might help:
by conditioning the critic on a chunk of actions rather than a single action, it adds information about the goal, producing better critic representations.

\vspace{-4pt}
 \paragraph{The Representational Benefit.} To test whether the success of CRL + AC stems solely from the explanations discussed in prior work, we construct a setting where none of them apply and ask whether the gains survive. On three offline datasets where CRL + AC clearly outperforms CRL, we distill the chunked critic into a single-action one: we replace its state-action encoder $\phi(s, a_{1:H})$ with a single-action encoder $\phi'(s, a)$ trained by expectile regression onto the chunked critic's values (DQC~\citep{liDecoupledQChunking2025}; see Appendix~\ref{app:dqc}), and extract a conventional single-step policy from it. This policy outputs and executes a single action, so it cannot benefit from any of the standard explanations: \textit{(1)} it cannot model non-Markovian behavior, \textit{(2)} unbiased $H$-step returns play no role, since CRL is not a temporal difference method, and \textit{(3)} neither exploration nor temporally extended exploitation applies, since the data is pre-collected and the policy executes one action at a time. All it inherits from the chunked critic are its representations. If the chunking benefit came entirely from the standard explanations, this policy should perform no better than standard CRL; if the benefit is representational, it should retain a positive gain, even if smaller than that of the full-chunk policy.

\looseness=-1 In practice, the distilled single-step policy outperforms standard CRL on all
three datasets for most of training, with relative gains ranging from ${\sim}30\%$ to ${\sim}100\%$ depending on the dataset (Figure~\ref{fig:representational_benefit}). Since none of the standard explanations applies here, the surviving gain must originate in the critic's representations — and because distillation can only lose information, it is a lower bound on their true benefit. This benefit is also distinct from those identified in prior work and compounds with them: the full-chunk policy further improves the distilled single-step one (Figure~\ref{fig:all_dqc_learning_curves}). We attribute this further gain to non-Markovian modeling, the only standard explanation still available in this setting.

\paragraph{Action Chunking Adds Information.}
Having observed the representational benefit, we ask what action chunking adds to critic representations. We hypothesize that a chunk gives the critic more information about the goal than a single action. To test this, we measure the critic's \emph{categorical accuracy}: for a batch of $B$ state-action/future state pairs, the critic scores all $B{\times}B$ pairings $\phi((s, a_{1:H})_i)^\top \psi(g_j)$, and accuracy is the fraction of states for which the true future $g_i$ scores highest among the $B$ candidates. Higher accuracy indicates that the critic's representation discriminates the correct future more sharply. To ensure any change reflects action information alone, we vary only the critic's action conditioning, holding the validation dataset, batch size $B$, and discount $\gamma$ fixed, so the number of negatives and the data distribution are unchanged. Crucially, we measure the action's contribution against a state-only critic (Eq.~\ref{eq:value_critic}): the state already carries most of the discriminative signal, so the question is what the action, and then the chunk, add on top of it.

Empirically, categorical accuracy increases monotonically with $H$ on the validation dataset (Figure~\ref{fig:why_ac_helps}, Left). Relative to the state-only critic, conditioning on a single action ($H{=}1$)
yields ${\sim}4.5$ percentage points higher accuracy, and conditioning on a chunk yields a larger gain, peaking at
${\sim}9.5$ percentage points at $H{=}30$, nearly double the single-action gain. These results support our hypothesis: conditioning on a chunk gives the critic more information about the goal than a single action, and the critic captures it. The exception is $H{=}50$, where validation accuracy drops even as \emph{training} accuracy peaks. This is overfitting, not a loss of information: the critic memorizes the training data instead of generalizing. Formally, the rise in accuracy with $H$ is consistent with the data-processing inequality: the mutual information between the state-action input and the goal can only increase with action chunk length, $I(s, a_{1:H}; g) \ge I(s, a; g)$.

\paragraph{Why Does Performance Collapse at Large $H$?}
Critic quality and policy success diverge at large $H$: validation categorical accuracy keeps rising with $H$, peaking at $H{=}30$, yet success peaks at $H{=}5$ (${\sim}72\%$) and falls to near zero by $H{=}30$ (Figure~\ref{fig:why_ac_helps}). If longer chunks yield better critics, why does performance collapse? We test whether the collapse is a failure of the critic or of policy extraction. Since execution is fixed to single-step replanning ($H_{\text{exec}}{=}1$), open-loop execution cannot be the cause, leaving policy extraction as the suspect: extracting a policy from a long-chunk critic requires the actor to model a distribution over a full length-$H$ chunk, which grows harder with $H$~\citep{liDecoupledQChunking2025}. If the critic is the problem, its information should be unusable; if extraction is the problem, a simpler policy should be able to recover it. We therefore apply the same DQC distillation used earlier to isolate the representational benefit, now with a moderate policy horizon ($H_\pi{=}5$) rather than a single action, applied to the $H{=}30$ and $H{=}50$ critics (details in Appendix~\ref{app:dqc}).

Empirically, this recovers success from near zero to ${\sim}50\%$ (Figure~\ref{fig:why_ac_helps}, Right, light bars). And since distillation can only lose information, never add it, then the recovered policy cannot be more informed than the original $H{=}30$ and $H{=}50$ critics. The fact that a policy extracted from the distilled short-chunk critic reaches ${\sim}50\%$ therefore shows the critics themselves held usable information; the undistilled collapse was a failure of policy extraction, not of critic quality. At large $H$, the bottleneck is extracting a policy from the critic, not the critic itself, and the representational benefit still holds.

\subsection{Ablation Studies}
\label{sec:exp_ablations}


\begin{figure}[t]
    \vspace{-12mm}
    \centering  \includegraphics[width=0.9\textwidth]{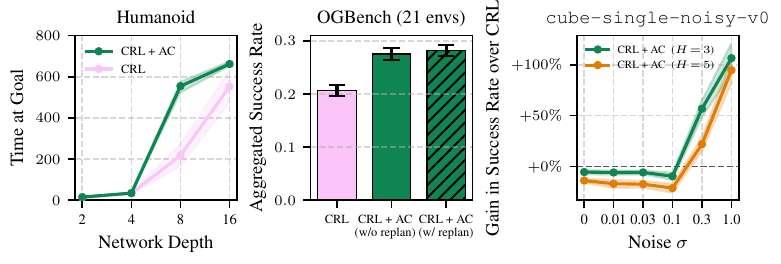}
     \vspace{-4mm}
    \caption{\textbf{Ablations on Action-Chunked CRL.}
    \textit{(Left)} \textbf{Performance across network depths.} We evaluate the effect of network depth scaling on a challenging Humanoid task, for CRL and CRL + AC with action chunk length $H {=} 3$. CRL + AC matches or exceeds CRL at every depth. \textit{(Middle)} \textbf{Effect of replanning on CRL + AC.} On 21 OGBench
    environments, we compare CRL + AC ($H{=}3$) executing the full chunk open-loop
    ($H_{\text{exec}}{=}H$) against replanning every step ($H_{\text{exec}}{=}1$),
    alongside standard CRL. Replanning every step achieves comparable average
    performance to open-loop execution, though its effect varies across environments
    (Appendix~\ref{app:replan}).
    \textit{(Right)} \textbf{Action-Chunked CRL and noise.} We report the success rate gain of CRL + AC over CRL for action chunk lengths $H \in \{3, 5\}$ across \texttt{cube-single-noisy-v0}. We vary the noise level $\sigma$ injected into expert actions during offline dataset collection, and train both CRL and CRL + AC on the resulting datasets. Gains become consistent for $\sigma {\geq} 0.3$, reaching up to $100\%$ at the highest noise level.}
\label{fig:ablations}
\vspace{-2mm}
\end{figure}

\paragraph{Network Depth Scaling.} Action chunking expands what the critic and actor can express; a natural question is whether the same benefit could be obtained simply by increasing network capacity. We therefore ablate network depth for CRL and CRL + AC, following the scaling setup of \citet{wang20261000} (details in Appendix~\ref{app:network_depth}). The left panel of Figure~\ref{fig:ablations} shows that on the Humanoid task the benefit of action
chunking persists as network depth increases: CRL + AC matches or exceeds CRL at every depth, so
its advantage compounds with capacity rather than being substituted by it. The effect is more
pronounced on Ant Hardest Maze (Appendix~\ref{app:network_depth}), where action chunking is
markedly more compute-efficient: a 2-layer CRL + AC network matches the performance of a
$8\times$ deeper (16-layer) CRL network; added depth is therefore not a substitute for action chunking.

\paragraph{Open-loop vs. Replanning.} By default we execute chunks open-loop, which conflates the representational effect with the effect of committing to a sequence of actions; we therefore ablate the evaluation replanning on 21 OGBench
environments with action chunk length $H{=}3$ (Figure~\ref{fig:ablations}, Middle). At evaluation we
execute only the first $H_{\text{exec}} {\le} H$ actions of each predicted chunk before requerying the
policy ($H_{\text{exec}}{=}H$ recovers full open-loop execution, $H_{\text{exec}}{=}1$ replans at
every step). Replanning every step is slightly beneficial offline, though it varies across environments, whereas online it degrades performance, falling below standard CRL (Appendix~\ref{app:replan}). Offline, open-loop execution is not the source of the gains. Online, the representational benefit and open-loop execution are entangled: the latter is necessary, so we cannot cleanly isolate the former. We expect the representational benefit to be present online as well, but leave disentangling the two to future work.

\paragraph{Varying Noise.} The offline results showed larger gains on noisy datasets; to test whether noise is the driver, we systematically vary the noise injected into expert actions during data collection on \texttt{cube-single-noisy-v0}, and then compare CRL and CRL + AC with $H \in \{3,5\}$ (Figure~\ref{fig:ablations}, Right). The advantage of CRL + AC grows almost monotonically with noise. The same trend holds on \texttt{puzzle-3x3-noisy-v0} (Appendix~\ref{app:varying_noise}), confirming the effect is not specific to one environment. Note that this experiment varies the dataset at a fixed chunk length, complementary to our earlier analysis. Our critic metrics (Figure~\ref{fig:vary_noise_appendix}) confirm action chunking adds information at every noise level, but this advantage stays roughly constant while the performance gap widens: noise amplifies the added information through a mechanism our metrics do not capture. Action chunking thus becomes more valuable as noise increases but the amplification mechanism remains open.

\begin{tcolorbox}[takeawaybox]
\begin{itemize}[leftmargin=16pt]
    \item \textbf{Action chunking consistently improves CRL.} CRL + AC improves over CRL across offline and online tasks ($+31.7\%$ offline manipulation, $+69.4\%$ on \texttt{noisy}/\texttt{explore}, $+111.2\%$ time-at-goal online) without degrading any task group.
    \item \textbf{Action chunking improves the critic by adding information.} An action chunk gives the critic more information about the goal than a single action, and the critic captures it: over a state-only critic, a chunk nearly doubles the validation accuracy gain of a single action. Distillation confirms that better critic representations contribute to the gain.
    \item \textbf{Critic quality and policy quality decouple at large $H$.} Beyond a moderate chunk length, a better critic does not mean a better policy: accuracy keeps rising to $H{=}30$ while success collapses (Figure~\ref{fig:why_ac_helps}). With execution held fixed, the bottleneck is \emph{policy extraction}, not critic quality. Distilling the long-chunk critic into a short-chunk one, and then extracting a policy, recovers performance from near zero to ${\sim}50\%$. The critic's information was usable all along.
    \item \textbf{Network scaling does not substitute for action chunking.} CRL + AC retains its advantage at every depth tested, and a shallow CRL + AC network can match a much deeper CRL one. Added depth is therefore not a substitute for action chunking.
\end{itemize}
\end{tcolorbox}

\section{Conclusions}
Treating CRL as a representative self-supervised method, we study the effect of action chunking beyond the settings where it has previously been studied, and show that conditioning the critic on short action chunks yields large gains across offline and online benchmarks. Investigating the source of these gains, we find that the explanations proposed for action chunking in other settings: modeling non-Markovian behavior, multi-step value estimation, and temporally extended execution do not fully account for its effectiveness here; instead, a mechanism specific to CRL is at play: conditioning on a chunk gives the critic more information about the goal than a single action, improving its representations and, in turn, the effectiveness of the contrastive critic.
We further characterize when this benefit saturates: very long chunks improve the critic but hinder policy extraction, and we show that decoupling the critic and policy horizons largely recovers performance. 

While we were able to observe consistent trends and significant gains, these remain conditional on the selection of the right chunk length.
In practice, we have found a mild horizon of three steps ($H{=}3$) to be generally the most beneficial; nevertheless, designing an offline procedure to automatically choose chunk lengths, possibly per state, is an important direction for future work. A natural next direction is whether the benefit extends to other temporal enrichments of the critic's input: action chunking, motivated by its established use in RL, is only one such method, and whether alternatives like conditioning on future states or past history yield similar gains remains an open question.
More generally, our work suggests that action chunking synergizes particularly well with CRL, and improves performance through mechanisms beyond those previously studied. Similar phenomena may arise beyond CRL: preliminary results in Appendix~\ref{app:other_ssrl} show that action chunking also improves other self-supervised RL methods, motivating its broader study in self-supervised RL.

\subsubsection*{Author Contributions}
Michal Korniak proposed studying action chunking in contrastive RL; ran the initial experiments establishing
its viability; designed, implemented, and ran the full offline experiments; designed the analysis of
why action chunking helps, including the use of DQC as a diagnostic; ran the varying-noise and
offline replanning experiments; wrote the initial drafts of the method, offline results, and analysis sections; and wrote the
initial outline of the paper. Kamil Dybek designed, implemented, and ran the full online experiments; ran the network-scaling and online replanning experiments; and wrote the initial draft of the online results section. Benjamin Eysenbach supervised the project; advised on the presentation of the work; and
contributed to writing and revising the manuscript. Marco Bagatella supervised the project;
advised on the experimental design and on the analysis of the synergy between CRL and action
chunking; and contributed to writing throughout the paper, particularly its positioning,
introduction, and discussion. Michał Bortkiewicz supervised the project; advised on the experimental design; and contributed to writing throughout the paper. 

\subsubsection*{Acknowledgments}
We gratefully acknowledge the Polish high-performance computing infrastructure PLGrid (HPC Center: ACK Cyfronet AGH) for providing computer facilities and support within the computational grant no. PLG/2025/018637.
Marco Bagatella is supported by the Max Planck ETH Center for Learning Systems. This project was supported in part by the Swiss National Science Foundation under NCCR Automation, grant agreement 51NF40 180545. Michał Bortkiewicz is supported by National Science Centre, Poland (grant no. 2023/51/D/ST6/01609). We thank Yarden As, Leander Diaz-Bone, and Alicja Ziarko for helpful feedback on the paper.

\bibliography{iclr2026_conference}
\bibliographystyle{iclr2026_conference}

\newpage
\appendix

\section{Extended Related Works}
\label{app:related}

\paragraph{Self-supervised Reinforcement Learning}
Contrastive reinforcement learning~\citep{eysenbach2022contrastive} is a prototypical self-supervised reinforcement learning algorithm, as it operates over datasets of trajectories with no reward labels.
As such, it fits within a broader family of self-supervised algorithms for representation and reinforcement learning, which we will now briefly discuss as they may also benefit from action chunking.
While CRL is contrastive and Monte-Carlo at heart, most self-supervised methods rely on either temporal-difference learning~\citep{touati2021learning, park2024foundationpolicieshilbertrepresentations}, on self-prediction~\citep{grill2020bootstrap, lawson2025self} or both~\citep{bagatella2025td}.
In some cases, these methods are designed for zero-shot reinforcement learning~\citep{touati2021learning, park2024foundationpolicieshilbertrepresentations, bagatella2025td}, i.e. for retrieving an optimal policy for arbitrary reward function; in others, they are mostly  targeting representation learning~\citep{grill2020bootstrap, lawson2025self} or dynamics modeling~\citep{maes2026leworldmodel}.
Despite their difference, these works are generally designed to be either action-independent~\citep{lawson2025self}, or to operate over single-step actions; due to their close connection to CRL, it is possible that they may benefit from action chunking through known and unknown mechanisms.

\section{Action-Chunked Contrastive Reinforcement Learning Details}

Algorithm~\ref{alg:ac-crl} shows the critic and actor losses, adapted from \citet{eysenbach2022contrastive}. The only modifications to standard CRL are highlighted: the state-action encoder takes a flattened action chunk of shape \texttt{(batch, H * action\_dim)} rather than a single action, and the policy outputs a full chunk. 

\begin{algorithm}[h]
\caption{Action-Chunked Contrastive Reinforcement Learning: the critic and actor losses.}
\begin{lstlisting}[language=Python, escapechar=@]
from jax.numpy import einsum, eye
from optax import  sigmoid_binary_cross_entropy
def critic_loss(states, action_chunks, future_states):
@\colorbox{green!20}{\texttt{\ \ \ \ \# action\_chunks: (batch\_dim, H * action\_dim) - flattened chunk}}@
    sa_repr = sa_encoder(states, action_chunks)  # (batch_dim, repr_dim)
    g_repr  = g_encoder(future_states)           # (batch_dim, repr_dim)
    logits  = einsum('ik,jk->ij', sa_repr, g_repr)
    return sigmoid_binary_cross_entropy(
                logits=logits, labels=eye(batch_size))
def actor_loss(states, goals):
@\colorbox{green!20}{\texttt{\ \ \ \ \# policy outputs full chunk (batch\_dim, H * action\_dim)}}@
    action_chunks = policy.sample(states, goal=goals)
    sa_repr = sa_encoder(states, action_chunks)  # (batch_dim, repr_dim)
    g_repr  = g_encoder(goals)                   # (batch_dim, repr_dim)
    logits  = einsum('ik,ik->i', sa_repr, g_repr)
    return -1.0 * logits
\end{lstlisting}
\label{alg:ac-crl}
\end{algorithm}

\section{Experimental Details}
\label{appendix}

\subsection{Hyperparameters}
\label{app:hyperparameters}

Hyperparameters for the offline OGBench and online JaxGCRL experiments are reported in Tables~\ref{tab:offline_hyperparameters} and~\ref{tab:online_hyperparameters}, respectively. We keep all hyperparameters unchanged from the original CRL implementation unless otherwise specified.

\begin{table}[!htbp]
\centering
\begin{minipage}{0.49\textwidth}
\centering
\scalebox{0.85}{
\begin{tabular}{lc}
\toprule
Hyperparameter & Value \\
\midrule
\texttt{train\_steps} & 1{,}000{,}000 \\
\texttt{eval\_episodes} & 20 \\
\texttt{lr} & $3 \times 10^{-4}$ \\
\texttt{batch\_size} & 1024 \\
\texttt{actor\_hidden\_dims} & $6 \times 512$ \\
\texttt{value\_hidden\_dims} & $6 \times 512$ \\
\texttt{latent\_dim} & 512 \\
\texttt{layer\_norm} & True \\
\texttt{discount} & 0.95 \\
\texttt{actor\_loss} & FQL \\
\texttt{alpha} & $\{1, 3, 10\}$ (tuned per env.) \\
\texttt{num\_flow\_steps} & 10 \\
\texttt{action\_chunk\_length} & depends on experiment \\
\texttt{replanning\_interval} & full action chunk length \\
\texttt{contrastive\_loss} & Sigmoid BCE \\
\texttt{value\_geom\_sample} & True \\
\texttt{value\_p\_trajgoal} & 1.0 \\
\texttt{num\_seeds} & 3 \\

\bottomrule
\end{tabular}
}
\caption{\textbf{OGBench Offline CRL + AC Hyperparameters.} Default values used across all experiments unless otherwise specified.}
\label{tab:offline_hyperparameters}
\end{minipage}
\hfill
\begin{minipage}{0.49\textwidth}
\centering
\scalebox{0.85}{
\begin{tabular}{lc}
\toprule
Hyperparameter & Value \\
\midrule
\texttt{num\_timesteps} & 60{,}000{,}000 \\
\texttt{max\_replay\_size} & 10{,}000 \\
\texttt{min\_replay\_size} & 1{,}000 \\
\texttt{episode\_length} & 1{,}000 \\
\texttt{unroll\_length} & 62 \\
\texttt{discount} & 0.99 \\
\texttt{num\_envs} & 512 \\
\texttt{batch\_size} & 256 \\
\texttt{lr} & $3 \times 10^{-4}$ \\
\texttt{contrastive\_loss} & InfoNCE \\
\texttt{energy\_function} & L2 \\
\texttt{logsumexp\_penalty} & 0.1 \\
\texttt{network\_depth} & 2 \\
\texttt{network\_width} & 256 \\
\texttt{representation\_dim} & 64 \\
\texttt{action\_chunk\_length} & depends on experiment \\
\texttt{replanning\_interval} & full action chunk length \\
\texttt{num\_seeds} & 5 \\
\bottomrule
\end{tabular}
}
\caption{\textbf{JaxGCRL Online CRL + AC Hyperparameters.} Default values used across all experiments unless otherwise specified.}
\label{tab:online_hyperparameters}
\end{minipage}
\end{table}

\subsection{Computational Resources}
All experiments were conducted on a single NVIDIA GH200. OGBench runs took approximately 3 hours per setting, i.e., with or without action chunking. JaxGCRL runs took between 15 minutes and 9 hours, depending on the experimental setting, with a total runtime of 472 hours.

\subsection{Figures}

\paragraph{Figure~\ref{fig:frontpage}.} This figure summarizes the
main results as three grouped aggregates. CRL + AC uses action chunk length $H{=}3$.
The left panel (\emph{OGBench manipulation}) aggregates the
18 manipulation environments (90 tasks); the middle panel (\emph{OGBench
noisy + explore}) aggregates the 12 \texttt{noisy}/\texttt{explore} environments (60 tasks); the
right panel (\emph{JaxGCRL}) aggregates 11 online locomotion and navigation environments. The
per-environment values underlying the two OGBench panels are the same runs reported in
Table~\ref{tab:ogbench_results_means}. We exclude the visual (pixel-observation) OGBench
environments, evaluating only on state-based observations. For the offline (OGBench) panels, CRL and CRL + AC are trained with a
flow-matching policy~\citep{lipman2022flow} and FQL~\citep{park2025flowqlearning} policy
extraction, with $\gamma{=}0.95$ and the per-environment $\alpha$ selected as described in
Section~\ref{sec:experiments}. For the online (JaxGCRL) panel, we use the default JaxGCRL
architecture, training protocol, and DDPG-style policy extraction. All hyperparameters are listed in Appendix~\ref{app:hyperparameters}, and are identical for CRL and CRL + AC except for the
chunked critic input and actor output. Aggregated success rate and 95\% confidence intervals are
computed with RLiable~\citep{agarwal2021deep} via stratified bootstrap over seeds $\times$ tasks
($N{=}3$ seeds per OGBench environment and $N{=}5$ seeds per JaxGCRL environment). For online tasks, success rate is measured at the end of
training. Reported group improvements ($31.7\%$, $69.4\%$, $93.1\%$) are relative gains of
CRL + AC over CRL on each aggregate. At evaluation, the policy executes the full predicted action chunk open-loop
before requerying ($H_{\text{exec}}{=}H$); no intermediate replanning is used. The corresponding per-environment learning curves are provided in
Appendix~\ref{app:ogbench_best_fql_learning_curves} (offline) and Appendix~\ref{app:jaxgcrl_learning_curves}
(online).

\paragraph{Figure~\ref{fig:learning_curves_10envs}.} This Figure demonstrates the learning curves for 10 JaxGCRL locomotion and navigation environments.
We report average time-at-goal and 95\% confidence intervals for standard CRL and CRL + AC with action chunk length $H \in \{3, 5, 10\}$.
Compared with success rate, time-at-goal provides a more informative measure of policy quality, as it rewards agents that both reach the goal quickly and remain there consistently~\citep{bortkiewicz2025accelerating}.
We use the default JaxGCRL architecture, training protocol, and DDPG-style policy extraction.
All hyperparameters are listed in Appendix~\ref{app:hyperparameters}.
Overall, CRL + AC with $H {=} 3$ achieves the strongest performance, yielding an improvement of $+111.2\%$ over standard CRL on average.
CRL + AC with $H {=} 5$ also outperforms CRL in most environments, although it rarely exceeds the performance of CRL + AC with $H {=} 3$.
In contrast, CRL + AC with $H {=} 10$ is frequently the weakest-performing variant among the evaluated methods.
Additional learning curves for the JaxGCRL environments are reported in Appendix~\ref{app:jaxgcrl_learning_curves}.

\paragraph{Table~\ref{tab:ogbench_results_means}.} All runs use action chunk
length $H{=}3$ and execute the full chunk open-loop at evaluation ($H_{\text{exec}}{=}H$). CRL and
CRL + AC both use a flow-matching policy with FQL policy extraction, $\gamma{=}0.95$, and per-environment
$\alpha \in \{1,3,10\}$ selected on validation; the two methods are identical except for the
chunked critic input and actor output. We report success rate at the end of training, averaged
over the environment's evaluation goals and $N{=}3$ seeds; group aggregates (italic rows) and
95\% CIs are computed with RLiable via stratified bootstrap over seeds $\times$ tasks. The mentioned
gains are relative improvements of the RLiable aggregate: $32\%$ over manipulation (all
\texttt{play}/\texttt{noisy} cube, puzzle, and scene groups) and $69\%$ over the suboptimal
\texttt{noisy}/\texttt{explore} variants. Grey columns are baselines reported by
\citet{park2025ogbench} for reference only, not re-run here, with CIs obtained by parametric
bootstrap from their published means and standard deviations; these are computed differently from
our seed$\times$task bootstrap and are therefore not directly comparable in width. Bold marks
entries within the 95\% CI of the better of CRL and CRL + AC (both bold when tied). Overall, the group
average improves, and gains concentrate on the noisier datasets, consistent with the ablation (Appendix~\ref{app:varying_noise}). We provide the full learning curves in Appendix~\ref{app:ogbench_best_fql_learning_curves}.

\paragraph{Figure~\ref{fig:why_ac_helps}.} This experiment uses
\texttt{cube-single-noisy-v0} with a high injected noise level ($\sigma{=}1.0$, versus the default
$0.1$) to obtain a more challenging dataset. All
methods execute a single action and replan at every step ($H_{\text{exec}}{=}1$), so execution is
held fixed across chunk lengths and cannot explain the differences. We use $\alpha{=}1.0$ for all
runs, the best value for CRL and for CRL + AC at $H{\in}\{3,5\}$ on this dataset. DQC is not tuned:
we set the expectile parameter $\kappa{=}0.9$ and distill into a policy critic of horizon
$H_\pi{=}5$, chosen because $H{=}5$ was the best-performing chunk length for CRL + AC. Categorical
accuracy is reported on held-out (validation) and training data; all other settings match the
offline OGBench configuration (Appendix~\ref{app:hyperparameters}). We expand on the DQC diagnostic and provide full learning curves in Appendix~\ref{app:dqc}.

\paragraph{Figure~\ref{fig:ablations}.} This Figure summarizes three ablations that probe when and why action chunking
helps; each is a condensed view of a fuller study reported in Appendix~\ref{app:aditional}.  The left panel scales network depth and shows that
action chunking's benefit compounds with capacity rather than being substituted by it
(results for all tested environments are included in Appendix~\ref{app:network_depth}, while the learning curves for this ablation study are reported in Appendix~\ref{app:scaling_learning_curves}).
The middle panel compares open-loop chunk execution ($H_{\text{exec}}{=}H$) against
replanning every step ($H_{\text{exec}}{=}1$), showing that replanning matches open-loop
execution on average in the offline setting (Appendix~\ref{app:replan}). We summarize each below. The
right panel varies the noise injected into the offline data and shows that the gain of CRL + AC
over CRL grows with noise (full sweep, additional environments, and the corresponding critic
metrics in Appendix~\ref{app:varying_noise}).

\subsection{Reproducibility} We provide all the experimental details, hyperparameters, evaluation protocols, and release the code to ensure reproducibility of the results. 

\newpage
\section{Evaluation Protocols}

\subsection{OGBench Evaluation Protocol}
We evaluate the policy every 50{,}000 training steps by rolling out 20 episodes across 5 tasks per environment (OGBench default), totaling 20 evaluation checkpoints over 1{,}000{,}000 training steps. We report the mean success rate aggregated over tasks, averaged over 3 seeds with 95\% bootstrapped confidence intervals computed via RLiable~\citep{agarwal2021deep}. All runs and seeds are included in the reported results.

\subsection{JaxGCRL Evaluation Protocol}
We evaluate the policy 65 times throughout training, including an initial evaluation before training begins, with evaluations spaced uniformly across the training run. For readability, the reported learning curves display only 10 evaluation points. We report the mean success rate aggregated over tasks, averaged over 5 seeds with 95\% bootstrapped confidence intervals computed via RLiable. All runs and seeds are included in the reported results.

\subsection{Benchmarks}
We provide a list of all environments we used in the conducted experiments. 
\paragraph{OGBench environments list:}

\begin{multicols}{2}
\begin{itemize}
    \item \texttt{cube-single-play-v0}
    \item \texttt{cube-double-play-v0}
    \item \texttt{cube-triple-play-v0}
    \item \texttt{cube-quadruple-play-v0}
    \item \texttt{cube-single-noisy-v0}
    \item \texttt{cube-double-noisy-v0}
    \item \texttt{cube-triple-noisy-v0}
    \item \texttt{cube-quadruple-noisy-v0}
    \item \texttt{puzzle-3x3-play-v0}
    \item \texttt{puzzle-4x4-play-v0}
    \item \texttt{puzzle-4x5-play-v0}
    \item \texttt{puzzle-4x6-play-v0}
    \item \texttt{puzzle-3x3-noisy-v0}
    \item \texttt{puzzle-4x4-noisy-v0}
    \item \texttt{puzzle-4x5-noisy-v0}
    \item \texttt{puzzle-4x6-noisy-v0}
    \item \texttt{scene-play-v0}
    \item \texttt{scene-noisy-v0}
    \item \texttt{antmaze-medium-explore-v0}
    \item \texttt{antmaze-large-explore-v0}
    \item \texttt{antmaze-teleport-explore-v0}
\end{itemize}
\end{multicols}

\paragraph{JaxGCRL environments list:}
\begin{multicols}{2}
\begin{itemize}
    \item Ant
    \item Ant U-Maze
    \item Ant Big Maze
    \item Ant Hardest Maze
    \item Ant Ball
    \item Ant Push
    \item Humanoid
    \item Humanoid U-Maze
    \item Humanoid Big Maze
    \item Humanoid Hardest Maze
    \item Cheetah
\end{itemize}
\end{multicols}

\newpage
\section{Additional Experiments}
\label{app:aditional}
\subsection{Analysis: Decoupling Critic and Policy Horizons (DQC)}
\label{app:dqc}

\begin{figure}[h]
    \centering
    \includegraphics[width=1.0\textwidth]{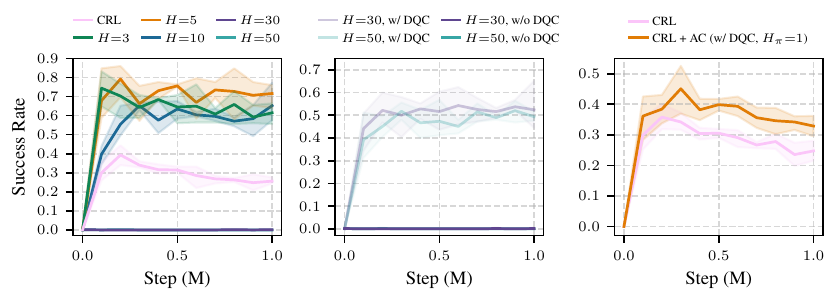}
    \caption{\textbf{Success over training across chunk lengths, and the DQC diagnostic.} Learning
curves on \texttt{cube-single-noisy} ($\sigma{=}1.0$), executing a single action with replanning at
every step ($H_{\text{exec}}{=}1$). \textit{(Left)} Success rate for CRL and CRL + AC across chunk
lengths $H \in \{3,5,10,30,50\}$. Moderate chunks improve substantially over CRL, peaking at
$H{=}5$, while $H{=}30$ and $H{=}50$ remain at ${\sim}0\%$ throughout training, showing the
large-$H$ collapse is present from the start of learning rather than a late-training instability.
\textit{(Middle)} DQC recovers performance at large $H$: the $H{=}30$ and $H{=}50$ critics stay at
${\sim}0\%$ without DQC but reach ${\sim}50\%$ when distilled into a short-chunk ($H_\pi{=}5$)
critic, showing the collapse is a policy-extraction failure, not a lack of critic information.
Shaded regions denote 95\% bootstrapped CIs. \textit{(Right)} Distilling the $H{=}5$ critic into a single-action ($H_\pi{=}1$) critic and extracting a one-step policy still outperforms standard CRL, despite conditioning on and executing a single action: indicating the representational benefit. The single-step policy does not fully match the chunked policy, however; the remaining gap can be attributed to the lossiness of distillation or to the chunked policy's ability to model non-Markovian behavior.}
    \label{fig:all_dqc_learning_curves}
\end{figure}

We use DQC~\citep{liDecoupledQChunking2025} in two ways: to isolate the representational benefit of action chunking by extracting a single-action policy from a chunked critic, and as a diagnostic to determine whether the performance
collapse at large $H$ (Subsection~\ref{subsec:why_does_ac_help}) stems from a poorly trained critic
or from the policy failing to extract from a well-trained one (Figure~\ref{fig:all_dqc_learning_curves}, Left). We jointly train the action-chunked critic with its standard contrastive loss and, alongside it, a short-chunk
state-action encoder $\phi'(s, a_{1:H_\pi})$ trained to match the long-chunk
critic's values by expectile regression:
\begin{equation}
    \mathcal{L}_\kappa(\phi') = \mathbb{E}\Big[\, \ell_\kappa\!\Big(
    \texttt{sg}\big[\phi(s, a_{1:H})^\top \psi(g)\big] - \phi'(s, a_{1:H_\pi})^\top \texttt{sg}\big[ \psi(g)\big]
    \Big)\Big],
\end{equation}
where $\ell_\kappa$ is the expectile loss with parameter $\kappa$ and $\texttt{sg}[\cdot]$ denotes
the stop-gradient, which prevents the distillation loss from affecting the long-chunk critic. The goal
encoder $\psi$ is shared and the long-chunk encoder $\phi$ is trained by the contrastive
objective; only $\phi'$ is trained by $\mathcal{L}_\kappa$. The policy is then extracted from $\phi'$ and $\psi$ using the same procedure as our base runs.
We use $(\kappa, H_\pi) = (0.9, 5)$ for the large-$H$ recovery and $(\kappa, H_\pi) = (0.95, 1)$
for the single-action distillation; all other hyperparameters match the main results.

\begin{figure}[h]
    \centering
    \includegraphics[width=1.0\textwidth]{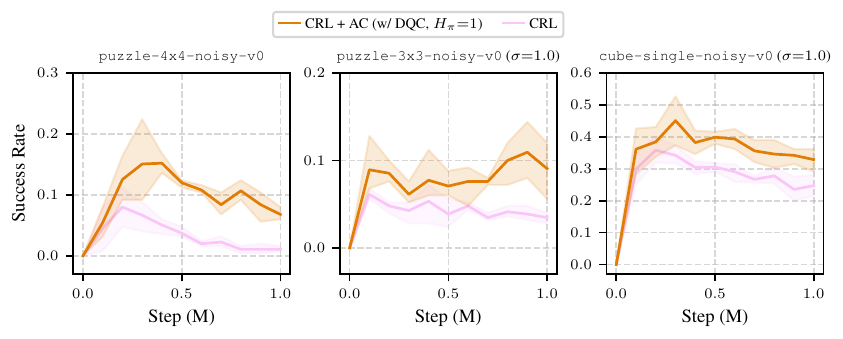}
    \caption{\textbf{The representational benefit survives distillation to a single-step policy.} On three datasets where CRL + AC performs well (\texttt{puzzle-4x4-noisy}, \texttt{puzzle-3x3-noisy} at $\sigma{=}1.0$, and \texttt{cube-single-noisy} at $\sigma{=}1.0$), we distill the CRL + AC with action chunk length $H{=}5$ critic into a single-action critic ($H_\pi{=}1$) via DQC and extract a conventional single-step policy (orange), then compare against standard CRL (pink). Despite outputting and executing only a single action, the distilled policy outperforms standard CRL throughout training on all three datasets, indicating that action chunking improves the critic's representation itself rather than acting solely through non-Markovian modeling, or temporally-extended execution/exploration. Shaded regions denote 95\% bootstrapped confidence intervals.}
    \label{fig:representational_benefit}
\end{figure}

\looseness = -1 We first apply this decoupling to a well-performing chunked critic. In Figure~\ref{fig:representational_benefit}, on three datasets where CRL + AC performs well we distill the $H{=}5$ critic into a single-action
($H_\pi{=}1$, $\kappa{=}0.95$) critic and extract a conventional one-step policy. This policy
conditions on and executes a single action, yet outperforms standard CRL throughout training.
Since the resulting policy is single-step, the gain cannot be attributed to modeling non-Markovian
behavior, unbiased $H$-step returns for value estimation, or temporally extended execution/exploration; it therefore isolates a
representational benefit of action chunking on the CRL critic itself (Subsection~\ref{subsec:why_does_ac_help}).

\newpage
The same decoupling serves a second purpose: recovering performance from critics whose chunks are too long to extract from directly. In the middle panel of Figure~\ref{fig:all_dqc_learning_curves}, the undistilled $H{=}30$ and $H{=}50$ critics remain at ${\sim}0\%$ success throughout training, whereas distilling each into a short-chunk ($H_\pi{=}5$, $\kappa{=}0.9$) critic and extracting a policy recovers success to ${\sim}50\%$ in both cases. Since the same critic yields ${\sim}0\%$ when extracted directly but ${\sim}50\%$ after distillation, its information was usable all along: the undistilled failure is one of policy extraction, not critic quality.

\newpage
\subsection{Varying Noise Ablation}
\label{app:varying_noise}
\begin{figure}[h]
    \centering
    \includegraphics[width=1.0\textwidth]{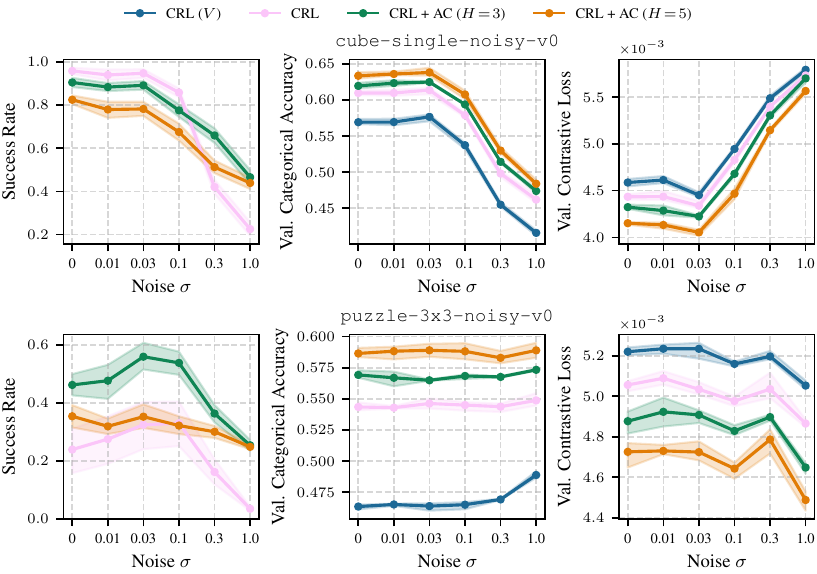}
    \caption{\textbf{Action chunking improves critic discrimination across noise levels.} Varying
    expert-action noise $\sigma$ during offline data collection on \texttt{cube-single-noisy-v0}
    (top) and \texttt{puzzle-3x3-noisy-v0} (bottom). Columns: task success, validation categorical
    accuracy, validation contrastive loss. Across both environments and all noise levels, accuracy
    increases and loss decreases from CRL ($V$) to CRL to CRL + AC ($H{=}3$) to CRL + AC ($H{=}5$),
    i.e.\ as more actions are included: reflecting more action information. CRL ($V$), which uses
    only the state, is already accurate, so the action adds only a few points, but they are the ones that matter for performance. Noise degrades discrimination sharply on cube but barely on puzzle, where the
    state stays informative. Shaded regions denote 95\% bootstrapped CIs.}
    \label{fig:vary_noise_appendix}
\end{figure}

We extend the main-body noise ablation
(Figure~\ref{fig:ablations}, Right) with two additional critic metrics across the noise sweep:
validation categorical accuracy and validation contrastive loss
(Figure~\ref{fig:vary_noise_appendix}). Across both environments and all noise levels, accuracy
increases and loss decreases as more actions are included in the critic input: from CRL ($V$),
which uses only the state, to CRL, to CRL + AC ($H{=}3$) and CRL + AC~($H{=}5$), confirming that
action chunking adds action information. The state-only critic CRL ($V$) is already accurate
($\sim\!57\%$ on cube), so the action contributes only a few points; but since the policy selects
actions by maximizing the critic at a fixed state, these few points are the ones that matter for
performance, which is why a small accuracy gain produces a large gain in success rate. Finally, the
effect of noise is environment-dependent: on cube it sharply degrades discrimination, giving
CRL + AC more room to improve, whereas on puzzle the metrics stay nearly flat: the state remains
informative regardless of action noise — yet chunking still improves performance. 

The state-only critic CRL ($V$) replaces the state--action encoder $\phi(s,a)$ with a state-only
encoder $\phi(s)$ in the contrastive objective, discarding the action:
\begin{equation}
\label{eq:value_critic}
    \mathcal{L}(\phi, \psi) = \mathbb{E}_{\substack{(s,a) \sim \mu_0 \\ g^+ \sim p^\pi(g \mid s, a) \\ g^- \sim p^\pi(g)}} \Big[ \log \sigma\!\left(\phi(s)^\top \psi(g^+)\right) + \log\!\left(1 - \sigma\!\left(\phi(s)^\top \psi(g^-)\right)\right) \Big].
\end{equation}

\subsection{Network Depth Ablation}
\label{app:network_depth}
For the network depth ablation, we adopt the network architecture design of
\citet{wang20261000}, using layer normalization after every dense layer and introducing skip
connections every 4 layers. We additionally increase the batch size to 512 and the training budget to $100{,}000{,}000$
environment steps on Ant Hardest Maze and $400{,}000{,}000$ environment steps on Humanoid. As in
the main paper (Figure~\ref{fig:ablations}, Middle), CRL + AC uses action chunk length $H{=}3$.

\begin{figure}[h]
    \centering
    \includegraphics[width=0.6\textwidth]{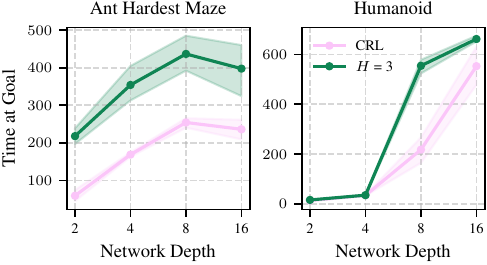}
    \caption{\textbf{Performance across network depths.} We evaluate the effect of network depth scaling on two challenging JaxGCRL tasks, Ant Hardest Maze and Humanoid, for CRL and CRL + AC with action chunk length $H = 3$. CRL + AC matches or exceeds CRL at every depth. On 2-layer network CRL + AC is comparable to the $8\times$ deeper network (16 layers) CRL, indicating better compute-efficiency. Shaded regions denote 95\% bootstrapped CIs.}
    \label{fig:network_depth}
\end{figure}

Figure~\ref{fig:network_depth} extends the network depth ablation from the main paper by adding
results on Ant Hardest Maze alongside Humanoid, sweeping network depth over $\{2, 4, 8, 16\}$.
CRL + AC outperforms or matches standard CRL at every depth on both tasks. On Humanoid the advantage of CRL + AC persists as network depth increases. On Ant Hardest Maze the advantage is roughly constant across depths but yields a large compute saving: a 2-layer CRL + AC network is comparable to the $8\times$ deeper CRL network (16 layers). In both cases the benefit of action chunking compounds with network
depth rather than being substituted by it. Full results, including additional chunk lengths and
 learning curves per-depth, are provided in Appendix~\ref{app:scaling_learning_curves}.

\newpage
\subsection{Replanning Interval Ablation}
\label{app:replan}
The replanning interval $H_{\text{exec}}$ denotes the number of actions executed before requerying the policy. We ablate this parameter in both the offline and online settings, comparing the execution of the whole action chunk ($H_{\text{exec}}{=}H$) with replanning after every action ($H_{\text{exec}}{=}1$). In the online setting, the replanning interval is shortened only during evaluation, and during experience collection, it remains equal to the action chunk length.

\begin{figure}[h]
    \centering
    \includegraphics[width=0.6\textwidth]{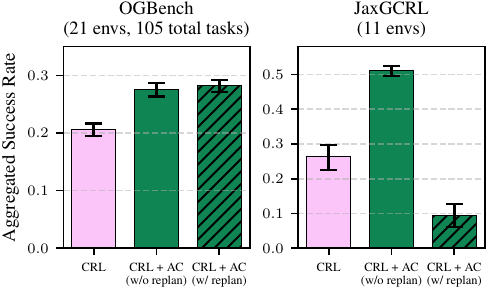}
    \caption{\textbf{Effect of replanning on action chunked CRL.} We compare the aggregated success rate of CRL + AC with replanning ($H_{\text{exec}}{=}1$) against CRL + AC without replanning ($H_{\text{exec}}{=}H$) and standard CRL across 2 benchmarks, reporting 95\% bootstrapped confidence intervals. The action chunk length $H$ is set to 3 for both CRL + AC variants. Overall, replanning does not significantly affect performance on the offline OGBench benchmark, but it is detrimental to performance on the online JaxGCRL benchmark.}
    \label{fig:replanning}
\end{figure}

Figure~\ref{fig:replanning} shows that reducing the replanning interval has no significant effect on the aggregated success rate in the offline setting, although the impact varies across environments (per-environment learning curves are provided in Appendix~\ref{app:ogbench_learning_curves_replanning}). In contrast, replanning after every action in the online setting substantially degrades performance, reducing it to well below that of standard CRL (learning curves are included in Appendix~\ref{app:jaxgcrl_learning_curves_replanning}).

Simultaneously leveraging the improved contrastive critics learned with action chunks while recovering the policy reactivity of closed-loop execution remains an open problem in the online setting. The cause of the stark contrast between the offline and online settings is also unclear. It may arise from several differences between these regimes, including their training protocols and evaluation environments. As test-time execution is orthogonal to our main contribution, namely the representational benefit of action chunking on the contrastive critic, we leave these questions to future work.

\newpage
\subsection{Discount Gamma Ablation}
\label{app:gamma_ablation}

\begin{figure}[h]
    \centering
    \includegraphics[width=1.0\textwidth]{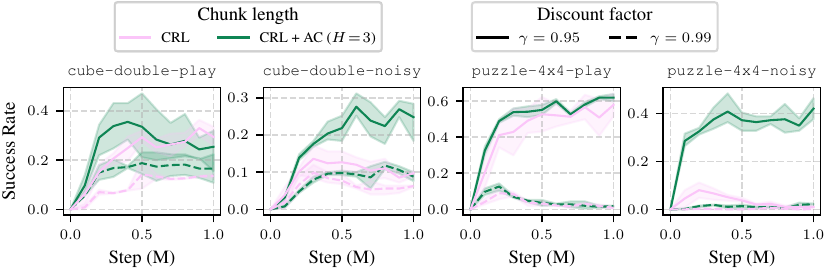}
    \caption{\textbf{Effect of discount factor $\gamma$ on CRL and CRL + AC.} We compare $\gamma = 0.95$ and $\gamma = 0.99$ for both CRL ($H = 1$) and CRL + AC ($H = 3$) across four representative OGBench tasks spanning two task types (Cube, Puzzle) and two dataset types (\texttt{play}, \texttt{noisy}). For each $\gamma$, the best-performing FQL $\alpha \in \{1, 3, 10\}$ is selected per environment and method. $\gamma = 0.95$ consistently outperforms $\gamma = 0.99$ for both methods across all settings, motivating our choice of $\gamma = 0.95$ throughout all experiments.}
    \label{fig:gamma_ablation}
\end{figure}

We ablate the discount factor $\gamma$ for both CRL and CRL + AC, comparing $\gamma = 0.95$ against the standard $\gamma = 0.99$. Results in Figure~\ref{fig:gamma_ablation} show that $\gamma = 0.95$ performs better across all evaluated tasks, and we therefore adopt it as the default in all experiments.

\newpage
\subsection{Action Chunking in Other Self-Supervised RL Methods}
\label{app:other_ssrl}

This paper studies the mechanisms through which action chunking improves CRL. CRL, however, is only one instance of self-supervised reinforcement learning, a broad family of methods that learn representations and behaviors from unlabeled interaction data (see Appendix~\ref{app:related} for a discussion of related work in this area). Action chunking may benefit these other methods as well, possibly through mechanisms that differ from the one we identify in CRL and that remain to be understood.

\begin{figure}[h]
    \centering
    \includegraphics[width=0.7\textwidth]{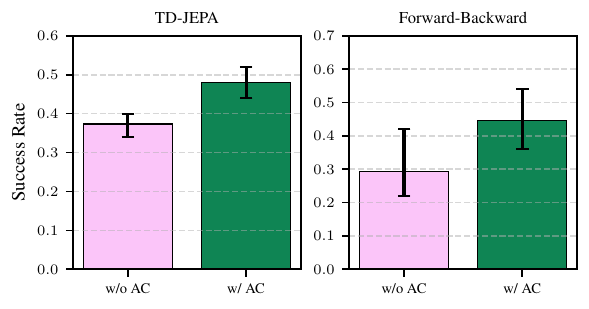}
    \caption{\textbf{Action chunking improves other self-supervised RL methods.} We report
success rate with and without action chunking (action chunk length $H{=}3$, executed open-loop)
for TD-JEPA and Forward-Backward on \texttt{cube-single-play-v0}. Action chunking raises
success rate by ${\sim}30\%$ on TD-JEPA and ${\sim}50\%$ on Forward-Backward. These
preliminary results suggest the benefit of action chunking extends beyond CRL, though we
do not isolate the source of improvement in these temporal-difference-based methods.
Error bars show 95\% bootstrapped confidence intervals.}
    \label{fig:other_ssrl}
\end{figure}

To encourage research in this direction, we report preliminary results of incorporating action chunking into two additional self-supervised methods, TD-JEPA~\citep{bagatella2025td} and Forward-Backward~\citep{touati2021learning}. We find that action chunking improves both methods (Figure~\ref{fig:other_ssrl}), raising success rate by ${\sim}30\%$ on TD-JEPA and ${\sim}50\%$ on Forward-Backward. We do not isolate the source of these gains here — unlike CRL, these methods are temporal-difference-based, so the multi-step-return explanation for action chunking applies and is entangled with any representational effect. Whether the representational mechanism we identify in CRL also drives the gains here is an open question, and a promising direction for future work.

\newpage
\section{Learning Curves}
\label{app:learning_curves}

\subsection{OGBench Best FQL}
\label{app:ogbench_best_fql_learning_curves}

\begin{figure}[h]
    \centering
    \includegraphics[width=1.0\textwidth]{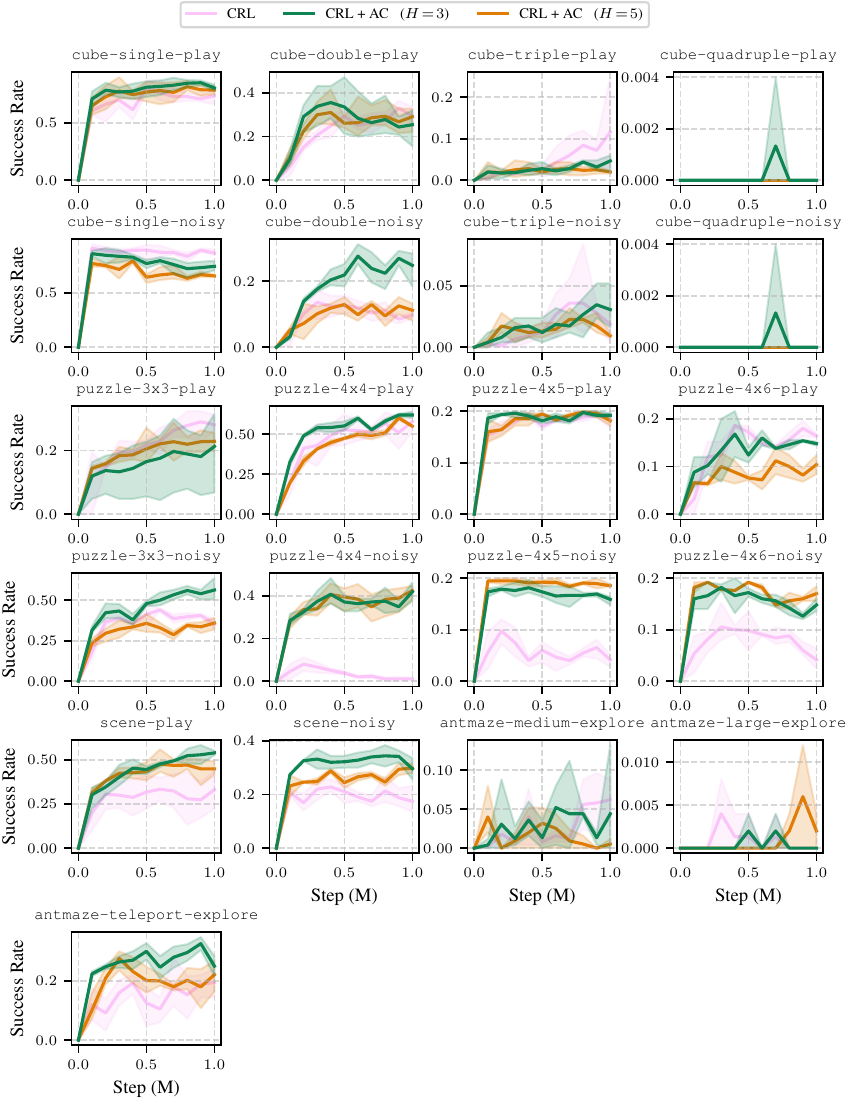}
    \caption{\textbf{Offline OGBench learning curves across all environments.} We present success rate learning curves for CRL and CRL + AC with $H \in \{1, 3, 5\}$ across  21 OGBench environments, with 95\% bootstrapped confidence intervals computed via RLiable. For each method and chunk length, the best-performing FQL $\alpha$ is selected per environment. CRL + AC consistently outperforms CRL across environments, with an average improvement of $32\%$.}
    \label{fig:best_fql_grid}
\end{figure}

\newpage
\subsection{OGBench Best FQL (Replanning)}
\label{app:ogbench_learning_curves_replanning}

\begin{figure}[h]
    \centering
    \includegraphics[width=1.0\textwidth]{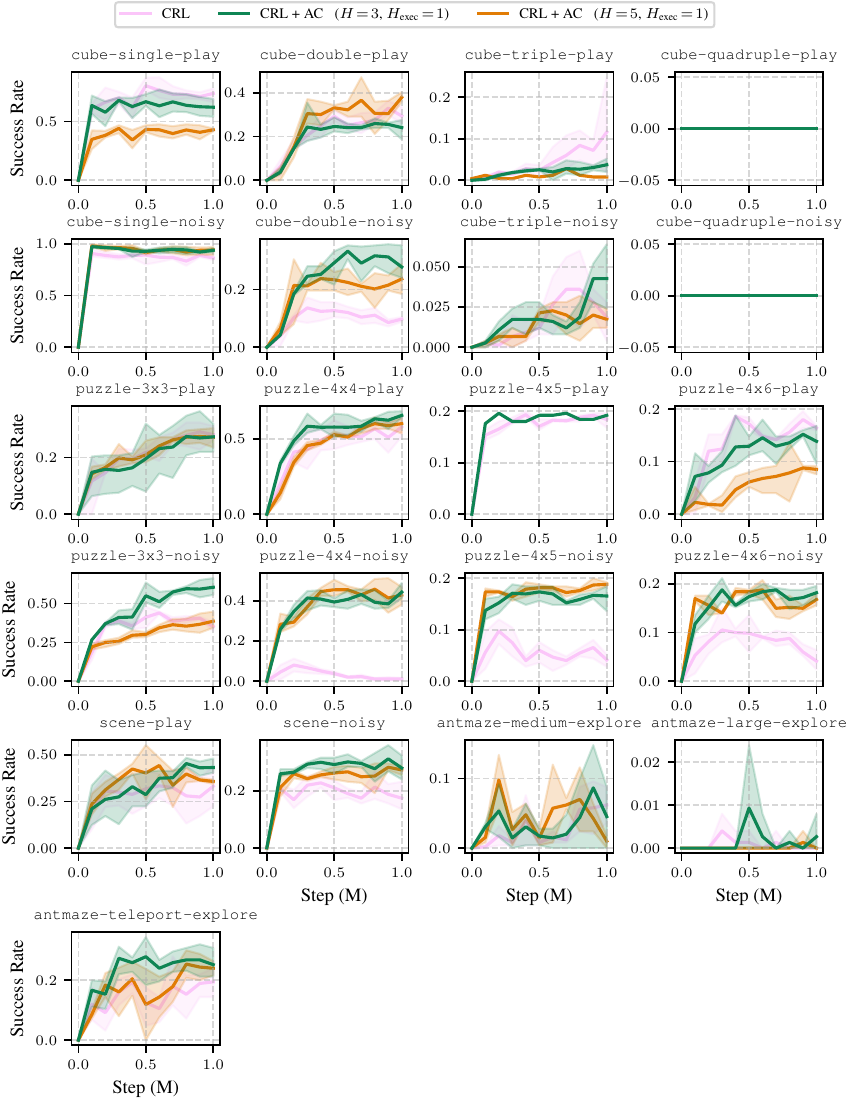}
    \caption{\textbf{Offline OGBench (Replanning) learning curves across all environments.} We present success rate learning curves for CRL and CRL + AC with $H \in \{1, 3, 5\}$ and replanning interval $H_{\text{exec}}=1$ across  21 OGBench environments, with 95\% bootstrapped confidence intervals computed via RLiable. For each method and chunk length, the best-performing FQL $\alpha$ is selected per environment. CRL + AC with replanning outperforms CRL across many environments, with an average improvement of $\sim$30\% — comparable to CRL + AC performance without replanning.}
    \label{fig:best_fql_grid_replan}
\end{figure}

\newpage
\subsection{OGBench All}

\begin{figure}[h]
    \centering
    \includegraphics[width=1.0\textwidth]{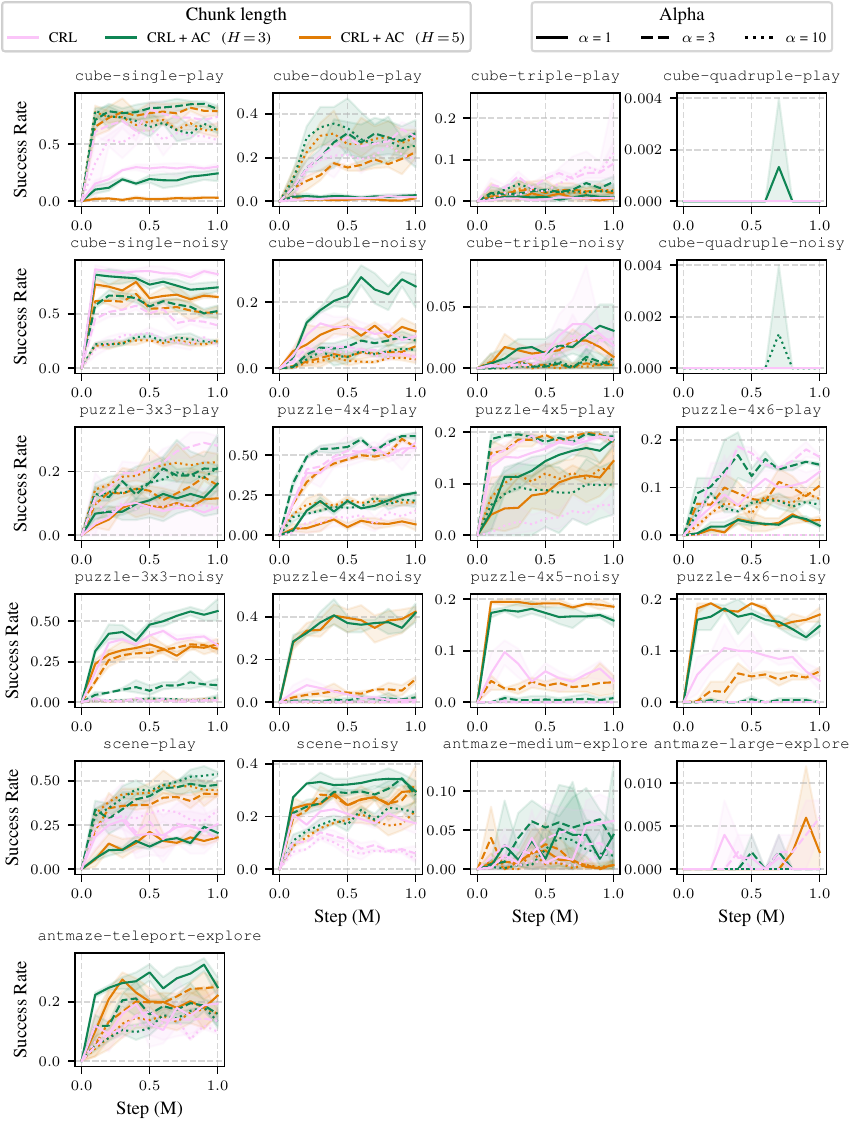}
    \caption{\textbf{Offline OGBench learning curves across all FQL $\alpha$ values.} We present success rate learning curves for CRL and CRL + AC with $H \in \{1, 3, 5\}$ across 21 OGBench environments for all evaluated FQL regularization strengths $\alpha \in \{1, 3, 10\}$, with 95\% bootstrapped confidence intervals computed via RLiable. In the main experiments we select the best-performing $\alpha$ per environment and method.}
    \label{fig:fql_grid}
\end{figure}

\newpage
\subsection{OGBench Varying Noise}
\begin{figure}[h]
    \centering
    \includegraphics[width=1.0\textwidth]{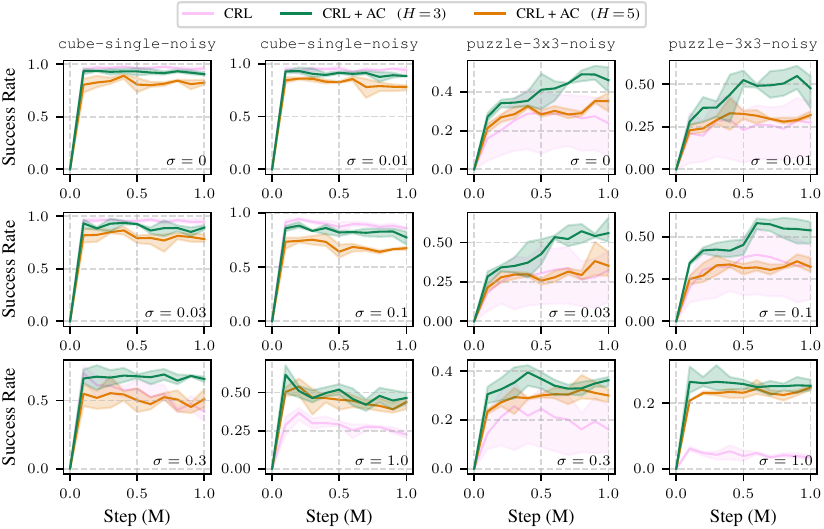}
    \caption{\textbf{Full learning curves for the varying noise experiment.} We present the complete learning curves corresponding to the experiment described in Figure~\ref{fig:vary_noise_appendix}, showing success rate over training steps for CRL and CRL + AC with $H \in \{3, 5\}$ across all noise levels $\sigma$. The performance gap between CRL + AC and CRL grows consistently with noise level, confirming that action chunking becomes increasingly beneficial as data quality degrades.}
    \label{fig:vary_noise_grid}
\end{figure}

\newpage
\subsection{JaxGCRL}
\label{app:jaxgcrl_learning_curves}

\subsubsection{Success Rate}
\begin{figure}[h]
    \centering
    \includegraphics[width=1.0\textwidth]{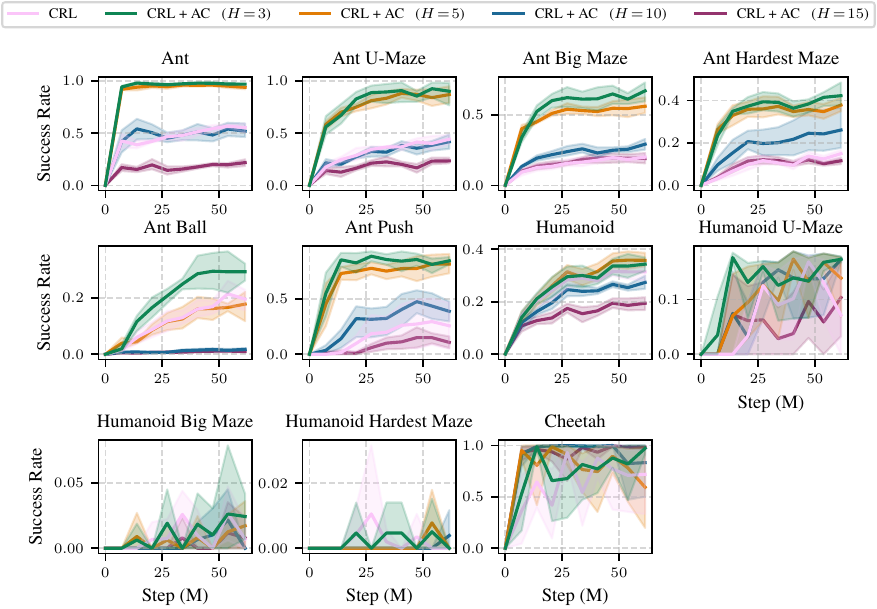}
    \caption{\textbf{Online JaxGCRL learning curves — success rate.} We present success rate learning curves for CRL and CRL + AC with $H \in \{3, 5, 10, 15\}$ across 11 JaxGCRL locomotion and navigation environments, with 95\% bootstrapped confidence intervals. CRL + AC with $H = 3$ is the best-performing variant overall, achieving approximately 90\% improvement over standard CRL on average.}
    \label{fig:jaxgcrl_grid_sr}
\end{figure}

\clearpage

\subsubsection{Time at Goal}
\begin{figure}[h]
    \centering
    \includegraphics[width=1.0\textwidth]{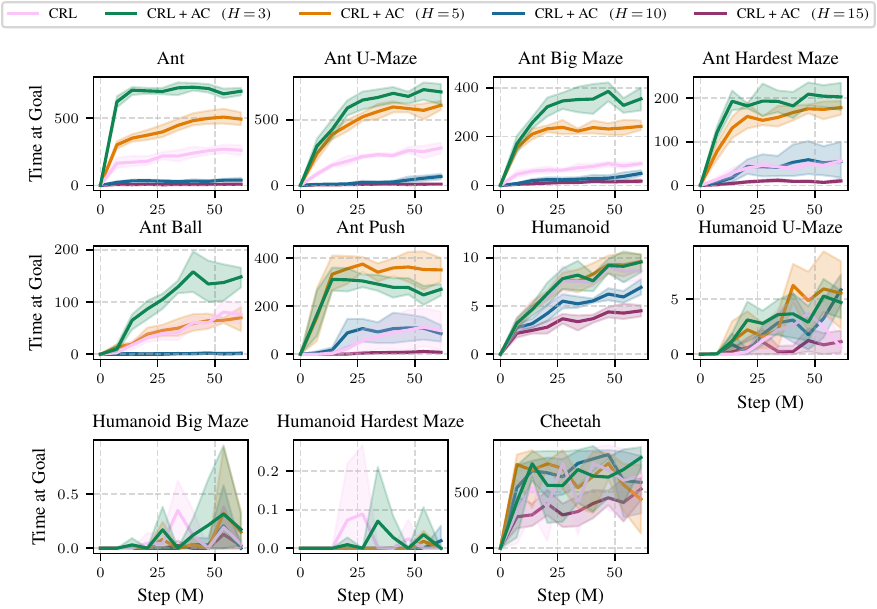}
    \caption{\textbf{Online JaxGCRL learning curves — time at goal.} We present time at goal learning curves for CRL and CRL + AC with $H \in \{3, 5, 10, 15\}$ across 11 JaxGCRL locomotion and navigation environments, with 95\% bootstrapped confidence intervals. Time at goal measures how long the agent spends at the goal state, rewarding both task success and behavioral stability~\citep{bortkiewicz2025accelerating}. CRL + AC with $H = 3$ is the best-performing variant, achieving $+111.2\%$ improvement over standard CRL on average.}
    \label{fig:jaxgcrl_grid_tag}
\end{figure}

\clearpage

\subsection{JaxGCRL (Replanning)}
\label{app:jaxgcrl_learning_curves_replanning}

\subsubsection{Success Rate}
\begin{figure}[h]
    \centering
    \includegraphics[width=1.0\textwidth]{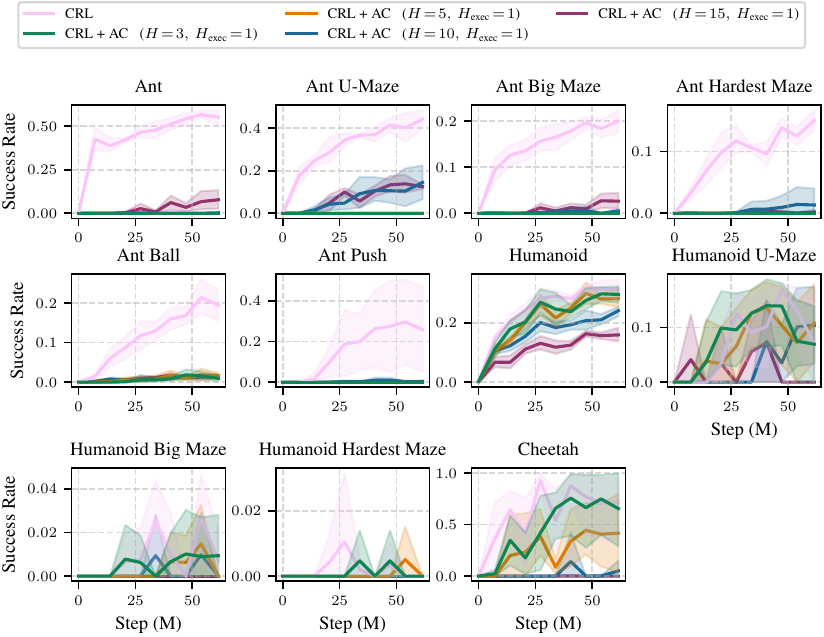}
    \caption{\textbf{Online JaxGCRL (Replanning) learning curves — success rate.} We present success rate learning curves for CRL and CRL + AC with $H \in \{3, 5, 10, 15\}$ and replanning interval $H_{\text{exec}}{=}1$ across 11 JaxGCRL locomotion and navigation environments, with 95\% bootstrapped confidence intervals. On these online tasks, single-step replanning ($H_{\text{exec}}{=}1$) underperforms
open-loop chunk execution (Appendix~\ref{app:replan}), in contrast to the offline setting; we leave
a full explanation of this offline-online difference to future work.}
    \label{fig:jaxgcrl_replanning_grid_sr}
\end{figure}

\clearpage

\subsubsection{Time at Goal}
\begin{figure}[h]
    \centering
    \includegraphics[width=1.0\textwidth]{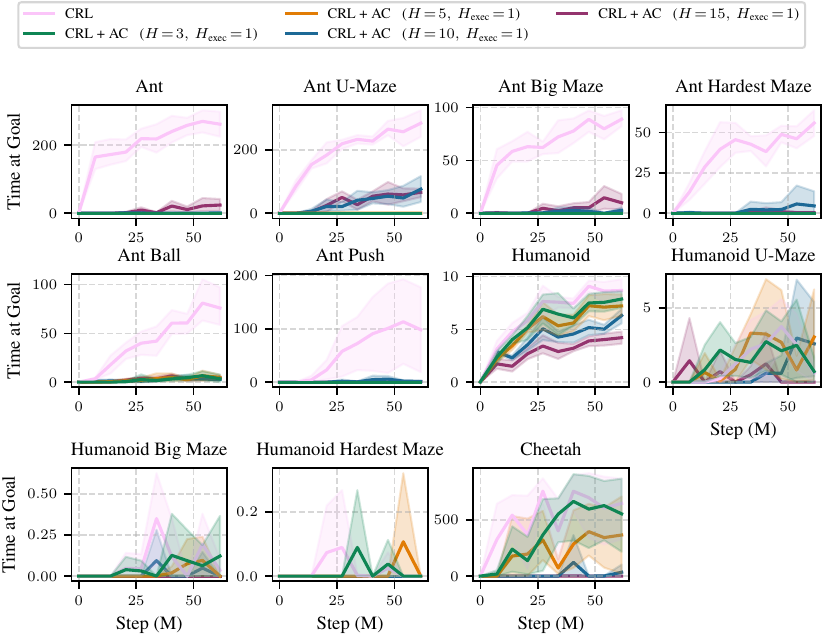}
    \caption{\textbf{Online JaxGCRL (Replanning) learning curves — time at goal.} We present time at goal learning curves for CRL and CRL + AC with $H \in \{3, 5, 10, 15\}$ and replanning interval $H_{\text{exec}}{=}1$ across 11 JaxGCRL locomotion and navigation environments, with 95\% bootstrapped confidence intervals. On these online tasks, single-step replanning ($H_{\text{exec}}{=}1$) underperforms
open-loop chunk execution (Appendix~\ref{app:replan}), in contrast to the offline setting; we leave
a full explanation of this offline-online difference to future work.}
    \label{fig:jaxgcrl_replanning_grid_tag}
\end{figure}

\clearpage

\subsection{JaxGCRL Network Depth Scaling}
\label{app:scaling_learning_curves}

\subsubsection{Success Rate}
\begin{figure}[h]
    \centering
    \includegraphics[width=1.0\textwidth]{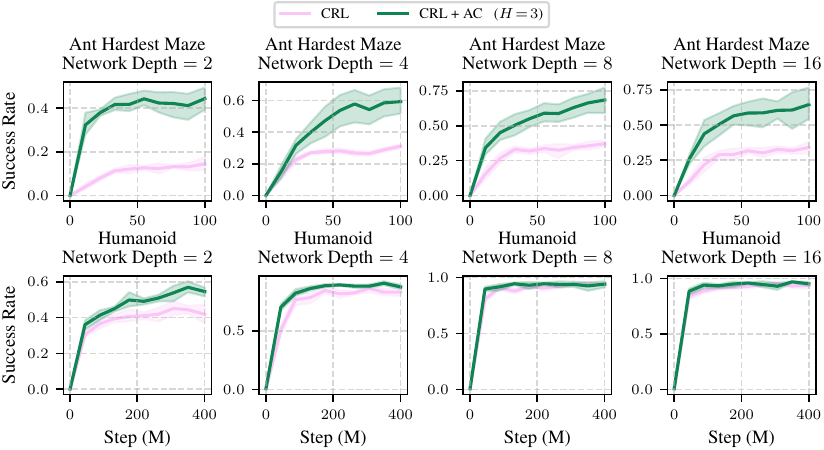}
    \caption{\textbf{Full learning curves for the network depth scaling experiment.} We present success rate learning curves for CRL and CRL + AC with action chunk length $H = 3$ across network depths on the Ant Hardest Maze and Humanoid JaxGCRL environments, with 95\% bootstrapped confidence intervals.}
    \label{fig:net_depth_grid_sr}
\end{figure}

\subsubsection{Time at Goal}
\begin{figure}[h]
    \centering
    \includegraphics[width=1.0\textwidth]{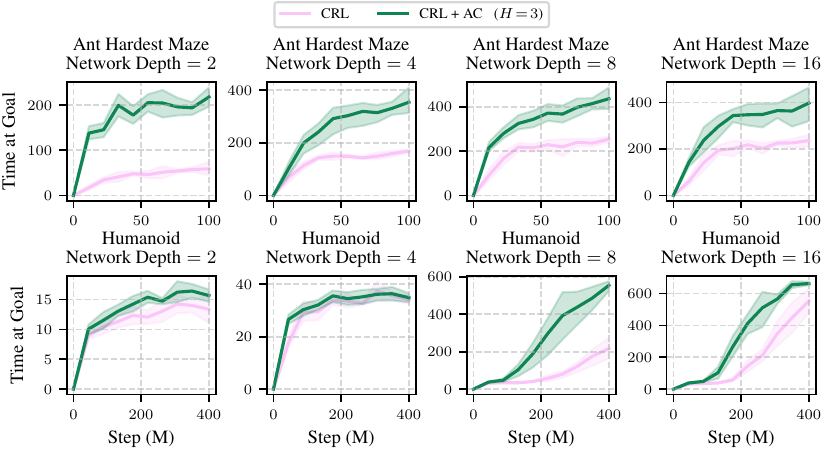}
    \caption{\textbf{Full learning curves for the network depth scaling experiment.} We present time at goal learning curves for CRL and CRL + AC with action chunk length $H = 3$ across network depths on the Ant Hardest Maze and Humanoid JaxGCRL environments, with 95\% bootstrapped confidence intervals.}
    \label{fig:net_depth_grid_tag}
\end{figure}

\end{document}